\documentclass{article}

\PassOptionsToPackage{numbers}{natbib}
\usepackage[preprint]{neurips_2023}
\usepackage[normalem]{ulem}

\usepackage[utf8]{inputenc} % allow utf-8 input
\usepackage[T1]{fontenc}    % use 8-bit T1 fonts
\usepackage{hyperref}       % hyperlinks
\usepackage{url}            % simple URL typesetting
\usepackage{booktabs}       % professional-quality tables
\usepackage{amsfonts}       % blackboard math symbols
\usepackage{nicefrac}       % compact symbols for 1/2, etc.
\usepackage{microtype}      % microtypography
\usepackage{xcolor}         % colors
\usepackage{makecell}

\hypersetup{
    colorlinks,
    linkcolor={red!50!black},
    citecolor={blue!50!black},
    urlcolor={blue!80!black}
}

\usepackage{caption}
\usepackage{subcaption}
\usepackage{algorithm}
\usepackage[noend]{algpseudocode}
\usepackage{amsmath,amssymb}
\usepackage{graphicx}
\usepackage{pifont}% http://ctan.org/pkg/pifont
\usepackage{array}
\usepackage{multirow}
\newcolumntype{L}[1]{>{\raggedright\let\newline\\\arraybackslash\hspace{0pt}}m{#1}}
\newcolumntype{C}[1]{>{\centering\let\newline\\\arraybackslash\hspace{0pt}}m{#1}}
\newcolumntype{R}[1]{>{\raggedleft\let\newline\\\arraybackslash\hspace{0pt}}m{#1}}

\definecolor{darkgreen}{rgb}{0.,0.65,0.2}

\newcommand{\cmark}{\ding{51}}%
\newcommand{\xmark}{\ding{55}}%

\title{CORAL: Contextual Response Retrievability Loss Function for Training Dialog Generation Models}

\author{%
  Bishal Santra\thanks{~~bsantraigi at gmail.com}\\
  CNeRG Lab, IIT Kharagpur\\
   \And
   Ravi Ghadia \\
   IIT Kharagpur \\
   \And
   Manish Gupta\thanks{~~gmanish at microsoft.com} \\
   Microsoft R\&D India \\
   \And
   Pawan Goyal \thanks{~~pawang at cse.iitkgp.ac.in}\\
   IIT Kharagpur \\
}

\begin{document}

\maketitle
% CORAL: A Contextualized Loss Function for Training Dialog Generation Models

\begin{abstract}
In the field of Natural Language Processing, there are many tasks that can be tackled effectively using the cross-entropy (CE) loss function. However, the task of dialog generation poses unique challenges for CE loss. This is because CE loss assumes that, for any given input, the only possible output is the one available as the ground truth in the training dataset. But, in dialog generation, there can be multiple valid responses (for a given context) that not only have different surface forms but can also be semantically different. Furthermore, CE loss computation for the dialog generation task does not take the input context into consideration and, hence, it grades the response irrespective of the context. To grade the generated response for qualities like relevance, engagingness, etc., the loss function should depend on both the context and the generated response. To address these limitations, this paper proposes CORAL,
% \footnote{The code for our research is publicly available \url{https://anonymous.4open.science/r/CORAL-Anonymous/}.}
a novel loss function based on a reinforcement learning (RL) view of the dialog generation task with a reward function that estimates human preference for generated responses while considering both the context and the response. Furthermore, to overcome challenges such as high sample complexity of RL training and a large action space, we propose a mix-policy training algorithm. Notably, using CORAL we can train dialog generation models without assuming the ground-truth as the only correct response. Extensive comparisons on benchmark datasets demonstrate that CORAL based models outperform strong state-of-the-art baseline models of different sizes.
% \pg{.. without assuming non-existence ..other than .... not a good phrasing. Can we mention the second point (based on context) above itself --- estimates human preference for generated response by ...} MG: Tried to handle this.}
%we can train dialog generation models without assuming non-existence of response other than the ground-truth, and CORAL loss is computed based on both the context and the response. 
% ACL 2023 Version of Contributions
% To address these limitations, in this paper, we propose a novel loss function, CORAL, that directly optimizes recent estimates of human preference for generated responses. Using CORAL, we can train dialog generation models without assuming non-existence of response other than the ground-truth. Additionally, the CORAL loss is computed based on both the context and the response. Extensive comparisons on two benchmark datasets show that CORAL based models outperform strong state-of-the-art baseline models of different sizes. 
% 

% -------------------------------------
\end{abstract}

\section{Introduction}\label{introduction}

Choosing the right loss function is crucial to get the expected behavior from deep learning based 
models trained for any task. While the 
token-level cross-entropy (CE) loss continues to excel in training natural 
language generation (NLG) models for various tasks, including dialog-response 
generation \citep{roller2020recipes, zhang2019dialogpt}, it is well accepted that CE is not the most appropriate choice of loss function for
training dialog generation models. Finding the right loss function for training  dialog generation is still an open problem and an active area of research \citep{shen2017conditional,zhao2017learning,saleh2020hierarchical,li2021improving}.

%Both unconditional or conditional NLG models are almost universally trained using
%the CE loss. 
The CE loss is computed by
comparing the predicted token probabilities to the ground truth target sequence
from the dataset. Thus, computation of CE loss is unconditional or
context-free, as it does not depend on the input prompt/context in the
case of conditional NLG tasks like dialog generation. 
To be able to generate responses with qualities like relevance, coherence, etc., 
the loss function should ideally consider both the context and the generated response.
% Without
% the context it is not possible to grade the generated response for qualities
% that human users truly care about, like relevance, coherence, etc.
%Another major concern with training dialog generation models using the CE 
%loss is the assumption of a fixed target. 
While training any NLG model using the CE loss,
the probability of the ground truth response is maximized. Here, we
make an implicit assumption that the ground truth is the only response
possible for the given context. This is a major concern as this property does not hold for most
dialogs where each context may have a large number of 
possible responses \citep{dou2021multitalk}\footnote{These problems are
specific to dialog systems only and may or may not apply to other NLG
tasks.}.

Previous attempts in training Seq2Seq dialog generation models using the
CE loss have led to various complications. Mode collapse is one of the
most common issues when training a Seq2Seq model with the CE loss,
mainly at smaller scales \citep{li2016deep,li2021improving}. Here, the model just ends up assigning a
high probability to one or more generic and bland responses e.g. ``I
don't know", ``I have a problem", ``Yes", etc., irrespective of the context.
% CE-based training of dialog generation models also have heavy data
% and model parameter-size requirements, which can be seen in successful
% CE-based models such as Blenderbot \citep{roller2020recipes} and DialoGPT
% \citep{zhang2019dialogpt}. These models have been pretrained on more than
% 1.5B and 147M dialog instances from, respectively. 
Previous research
works have also explored various augmentations to the model architecture \citep{serban2016building,zhao2019rethinking}
and/or the loss function \citep{serban2017hierarchical,shen2017conditional,zhao2017learning,li2021improving} to resolve the common problems with CE, as
mentioned above.

To train dialog generation models that maximize some user-expected qualities, we propose to directly optimize 
an estimate of human perceived quality of a context-response 
pair. \citet{Sinha2020LearningAU,yeh2021comprehensive} have shown that the output score of a dialog response retrieval model (trained on the same domain) correlates strongly with human perception of dialog response quality. 
Further,~\citet{santra2021representation} showed that representations learned by a
response retrieval model (using binary cross-entropy or a contrastive loss) capture
important dialog understanding features even better than large-scale dialog generation models trained using CE. 
% But that is not applicable for a generative setting.

\begin{figure}
    \centering
    \includegraphics[width=0.9\textwidth]{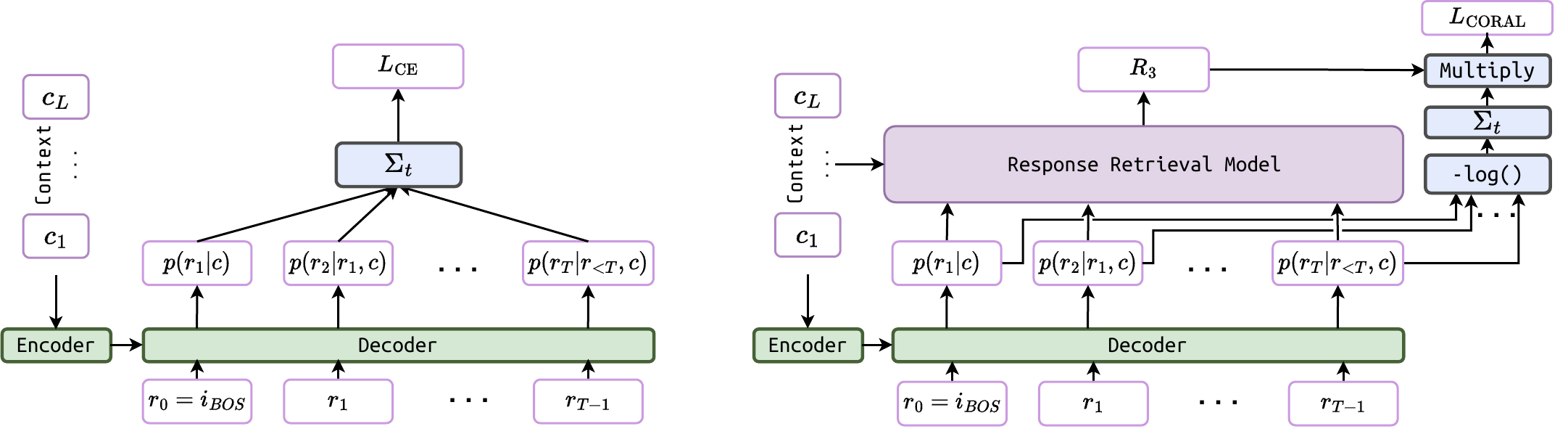}
    %https://drive.google.com/file/d/1naTj7ccr0aH8cVb5e1ZGIsRCJPyIa32P/view?usp=sharing
    \caption{Schematic of CE and CORAL losses (training time): CORAL optimizes a measure of compatibility, the $R_3$ reward (based on response retrieval models), between the context and a candidate response. 
    Compared to CORAL, CE is more strict and relies only on ground truth response targets for training. CORAL utilizes responses sampled from the trained network and updates its decoding-probability based on its $R_3$ reward value. 
    We utilize transformer-Seq2Seq models for our dialog generation experiments. 
    The context input is represented as $[c_1, ..., c_L]$, and the target response is denoted as $r = [ r_1, ..., r_T]$ which can either be the ground truth or a sample from the network.
    } %The encoder part of the Seq2Seq architecture has been hidden here to focus on the relevant part only.
    %For our implementation, we have used a response retrieval model for measuring this compatibility. 
    % The context input to the encoder is denoted by a token sequence $[c_1, c_2, ..., c_L]$, and the target response (ground truth or sampled from the network) sequence as $r = [ r_1,  r_2, ..., r_T]$.
    \label{fig:compare-losses}
\end{figure}

% These observations triggered us to propose a novel \textbf{Contextual Response Retrievability} loss function or \textbf{CORAL} that circumvents issues with CE loss and relies on a response retrieval model.  
% Further, we leverage CORAL to design a learning framework for dialog generation models that treats score calculated by retrieval models as a reward function, referred to as \textbf{Response Retrievability Reward} ($R_3$). 

Motivated by these findings, we introduce a novel loss function called Contextual Response Retrievability (CORAL) for training dialog generation models.
% we propose a novel \textbf{Contextual Response Retrievability} loss function or \textbf{CORAL}, for training dialog generation models, 
This loss function optimizes a response-retrieval model's score, referred to as the Response Retrievability Reward ($R_3$), as shown in Fig.~\ref{fig:compare-losses}. 
Our proposed learning-framework uses reinforcement learning (RL) to optimize the $R_3$ reward between context and generated responses. % using the trained policy. 
% \footnote{Responses generated from the trained model is directly used for training.}
To evaluate the effectiveness of CORAL, we train transformer-based Seq2Seq models \citep{vaswani2017attention} using the 
% cleaned version of 
DailyDialog dataset \citep{li2017dailydialog,wen-etal-2022-ddcc} for open-domain dialog generation and the DSTC7-Ubuntu dataset \citep{yoshino2019dialog} for domain-specific dialog generation.
We compare the performance against state-of-the-art CE-based Seq2Seq models and various other 
baselines using both automatic metrics and through human evaluation. In summary, our contributions are: 
  % (1) We propose CORAL loss function for training dialog generation models by directly optimizing for an estimate of human preference of a $\langle$context,\;response$\rangle$ pair. To the best of our knowledge, we are the first to propose a loss function for dialog generation model that also relies upon the context.
  %   (2) Further, we propose a recipe to train improved seq2seq dialog models which uses the CORAL loss in a reinforcement learning setup.
  % (3) We experimentally prove the effectiveness of CORAL against strong baseline models using CE or its variants. We make the code publicly available\footnote{\url{https://anonymous.4open.science/r/2022-CORAL-Anonymous/}}.
(1) the proposal of the CORAL loss function that directly optimizes for an estimate of human preference of a $\langle$context, response$\rangle$ pair, which is the first of its kind to also rely on the context, 
(2) a recipe for training improved seq2seq dialog models using the CORAL loss in a reinforcement learning setup, utilizing both on-policy and off-policy response samples, and 
(3) experimental evidence of the effectiveness of CORAL against strong baseline models using CE or its variants. 
% EXPIRED: https://anonymous.4open.science/r/2022-CORAL-Anonymous/

\section{Literature Review}\label{literature-review}

Inspired by the initial success of Sequence-to-Sequence (Seq2Seq) models in the machine translation task, \citet{ritter2011data} proposed a similar approach for training open-domain dialog (chitchat) generation models, using the CE loss. However, this vanilla approach suffered multiple shortcomings, such as token repetition in output generations, generic/bland responses that are ignorant of the context, etc. To remedy these shortcomings of the vanilla Seq2Seq architecture for the dialog generation task, various solutions have been proposed over the years. \citet{serban2016building,santra2021hierarchical} proposed methods for learning hierarchical representations of the context.  \citet{serban2017hierarchical,shen2017conditional,zhao2017learning,bao2019plato} developed latent variable models to capture the stochastic nature of the task. Recently, a significant amount of focus has been on pretraining large dialog generation models \citep{zhang2019dialogpt,bao2019plato,roller2020recipes,adiwardana2020towards} using the transformer architecture, following its recent unmatched success in almost every domain. Another set of important directions in dialog generation involve the development of response retrieval/next-utterance selection models \cite{lowe2015ubuntu,chen2019sequential,whang2021response,xu2021learning,humeaupoly,santra2021representation,henderson2020convert} and retrieval augmented generation or RAG \citep{wu2019response,gupta2020controlling,cai2021exemplar,komeili2021internet,zhu2018retrieval,komeili2021internet,guu2020retrieval} models. RAG models use the power of response retrieval models to select plausible responses/related knowledge from a large pre-existing corpus, and generate a response based on that. A significant difference between RAG and our approach is that we use the response-retrieval model as an optimization objective during training and do not use any external data source or knowledge-bases. %\pg{at inference time ?}} \bs{true for both training and inference}

Cross-entropy loss, while a popular choice for training dialog generation models, has its limitations in that it can only evaluate the generated response with respect to the ground truth response. To overcome this, various reinforcement learning (RL) based training algorithms have been proposed for training dialog generation models, which allow for optimizing other qualities of the model and output generation, e.g., coherence of the response. \citet{li2016deep} employed RL to optimize for long-term success, while \citet{li2017adversarial} incorporated adversarial learning and policy-gradient to create a discriminator-based reward function. Additionally, \citet{sankar2019deep,zhao2019rethinking,saleh2020hierarchical} utilized RL to learn latent discrete-action spaces for the purpose of training interpretable dialog generation models.

% \citet{li2016deep} used RL to optimize for long-term success in dialog generation models, \citet{li2017adversarial} combined adversarial learning and policy-gradient to model a discriminator-based reward function. \citet{sankar2019deep,zhao2019rethinking,saleh2020hierarchical} used RL to learn latent discrete-action spaces for training interpretable dialog generation models.
% , \citet{saleh2020hierarchical} applies a hierarchical reinforcement learning framework to learn abstract action spaces for dialog response generation.

In this paper, we propose a novel reward function based on response retrieval models for training RL-based dialog models. \citet{Sinha2020LearningAU} recently demonstrated that scores from retrieval models 
\cite{chen2019sequential,henderson2019convert,humeau2019poly}
correlate strongly with human judgments for model-generated responses. Hence, this paper explores the possibility of using this ``retrievability score'' as an optimization target and thoroughly analyzes it under different experimental settings to understand its effectiveness for dialog generation.

\section{Proposed Approach}\label{sec:methods}
% 2023May: SECTION TITLE CHANGED
% \enter{
Our proposed approach focuses on training a dialog generation model in a reinforcement learning (RL) setting by optimizing a novel reward function based on human preference estimates. To design the \textit{reward function}, which we call $R_3$, we use a response retrieval model as explained in Section~\ref{sec:R3reward}. We then propose a novel loss function, CORAL, which is based on this RL view of the dialog generation task to maximize the $R_3$ reward function, in Section~\ref{sec:formulation-of-the-coral-loss}. We use the CORAL loss to train a transformer-based Seq2Seq model instead of the traditional cross-entropy loss, as illustrated in Fig.~\ref{fig:compare-losses}. 
% formulate the dialog generation task as a RL problem and propose a REINFORCE 
% where we design a novel loss function, CORAL, based on the $R_3$ reward function.
% in Section~\ref{sec:formulation-of-the-coral-loss}, and train a transformer-based Seq2Seq model that optimizes \mg{the CORAL loss based on} the $R_3$ reward function instead of using the traditional cross-entropy loss, \mg{as illustrated in Fig.~\ref{fig:compare-losses}}. 
Furthermore, to overcome challenges such as high sample complexity of RL training and a large action space, we propose a \textit{mix-policy training algorithm} in Section~\ref{sec:training-policy}. 
% \enter{Our approach effectively trains the dialog generation model and improves its ability to generate high-quality responses. With this approach, we aim to create a dialog generation system that can mimic human-like responses in natural language conversations.}
Our approach effectively trains the dialog generation model, enhancing its ability to generate high-quality, human-like responses in natural language conversations.

% \pg{Can we refer to Figure 1 here?}\mg{Done}

\subsection{$R_3$ Reward Function}
\label{sec:R3reward}
% \enter{
Although cross-entropy loss is the norm for training Seq2Seq models on various tasks, 
%\mg{it is not specifically designed for dialog generation modeling. \sout{
it does not correlate with any quality that we expect a dialog generation model to possess \cite{liu2016not}. So, a natural research question is how we can design an alternative objective function that might be able to improve qualities like relevance, coherence, topicality, engaging etc. in a dialog system. 

The task of response retrieval or next utterance selection \cite{lowe2015ubuntu} involves predicting whether, given a dialog context and a candidate response, the response is a valid continuation to the context. \citet{Sinha2020LearningAU} showed that output probabilities from a model trained with tasks akin to response retrieval correlate strongly with human preferences for dialog responses, 
which means that the output (a value between $0$ and $1$) indicates whether a human annotator would rate the response as a coherent and on-topic continuation to the context. 
Based on this idea, we train a response retrieval model (on a given dataset) and define the output probability from the \textit{finetuned} retrieval model as the reward function to be optimized by a dialog generation model. We refer to this output probability score from a response retrieval model as the contextual \textbf{Response Retrievability Reward ($R_3$)}.

\paragraph{Training the Reward Model}
% \change{Should we add a figure for reward model training?}
% \enter{
We experiment with two model architectures for designing the reward function: ESIM (Enhanced Sequential Inference Model)~\cite{chen2019sequential} and BERT (Bidirectional Encoder Representations from Transformers)~\cite{devlin2018bert}. ESIM is an LSTM and cross-attention based model for sentence-pair classification. For further details about the ESIM architecture, please refer to the works by \citet{chen2016enhanced,chen2019sequential}. The BERT-based response retrieval model is designed by adding a classification layer on top of the CLS output representation (final-layer) and then finetuning the whole model for the response retrieval task.

% \enter{
The response retrieval model is trained in a self supervised setting using the binary cross-entropy (BCE) loss. The data for training the model is gathered as follows. 
% We train the response retrieval model using the binary cross-entropy (BCE) loss and a \emph{self-supervised} task defined as follows. 
We create positive and negative $\langle$context, response$\rangle$ pairs for this task from the target dialog dataset itself (same as the one to be used for dialog generation training).
% We create data samples for this task from the target dialog dataset (same as the one to be used for dialog generation training), we create positive and negative pairs of context and responses. 
For creating the positive samples, we extract context-response (CR) pairs by unrolling\footnote{If (A, B, C, D, E, F, G) is the sequence of utterances in a dialog, then it can be unrolled to create multiple CR-pairs, such as, $\langle$(A),B$\rangle$, $\langle$(A, B),C$\rangle$, $\langle$(A, B, C),D$\rangle$, and so on.} dialogs from the dataset. Then for each positive CR-pairs, we generate $n(=4)$ negative CR-pairs by pairing the context with random utterances from other dialogs in the dataset.

% \noindent\textbf{Contextual Response Retrievability Reward ($R_3$)}\label{sec:response-retrievability-reward} We define the output score of an already trained response retrieval model for a context-response pair as the Contextual Response Retrievability Reward. This score is usually normalized between $0$ and $1$ and indicates the probability that the response is a valid (coherent and on-topic) continuation to the context. \change{Could you please motivate why ESIM and BERT-based $R_3$ reward validate coherence and on-topic?}
% In general, retrieval models compute this score based on both the input context and the output response from a human annotator or a generative model.
% We refer to this score as a reward
% as it is not differentiable with respect to the input to the
% response retrieval model. 
% This input is usually a tokenized text sampled
% from a decoder or directly obtained from the dataset. 

\subsection{CORAL Loss Function}
\label{sec:formulation-of-the-coral-loss}
To be able to optimize the proposed $R_3$ reward function, which is not differentiable w.r.t. the policy network (implemented as a Seq2Seq model) parameters, we pose the response generation problem as a reinforcement learning (RL) task. Then, we apply REINFORCE~\cite{williams1992simple} to obtain a differentiable objective function, as described next.
% \footnote{As the reward function would not be differentiable with respect to the parameters of the policy network (implemented using a Seq2Seq-model), we first formulate response generation as an RL task. Then, we apply REINFORCE~\cite{williams1992simple} to obtain a differentiable objective function, as described next.} 
Each instance of the context-to-response generation task is considered as an episode in the RL formulation. The state consists of the tokenized dialog context and the set of response tokens generated until the previous timestep. Each episode consists of several actions taken by the agent, in our case the decoder, spanning the generation of a complete response. Each action corresponds to generation of an output token. The episode ends when the agent generates an EOS (end-of-sequence) token or has produced a max number of allowed tokens ($T$). Then the response retrieval model generates the $R_3$ reward, for the $\langle$context $c$, generated response $r\rangle$ pair.

The updates to the weights $\theta$ of the Seq2Seq (S2S) model $P(r|c)$, that maximize the expected return, $\mathop{\mathbb{E}}_{r\sim P(r|c)}[R_3(r,c)]$ are determined by the Episodic REINFORCE
algorithm~\cite{williams1992simple} %\footnote{See Section 5 in \cite{williams1992simple}} 
as follows.
\setlength{\abovedisplayskip}{4pt}
\setlength{\belowdisplayskip}{4pt}
\begin{align}
  \Delta \theta = \eta R_3(c,r) \sum_{t=1}^{T} \frac{\partial\log P(r_t|r_{<t},c)}{\partial \theta}
  \label{eqn:coral-update}
\end{align}
\noindent where $\eta$ is the learning rate, and $r$ is a response sampled from the learned policy $P(r|c)$.
% \textcolor{red}{MG: $\delta$? you meant $\partial$}
Loss function to be minimized (for an autoregressive decoder) can then be written as follows:
\begin{align}
    L_{\text{CORAL}}=-R_3(c,r) \sum_{t=1}^{T} \log P(r_t|r_{<t},c)
    \label{eqn:loss-coral}
\end{align}
% \textcolor{red}{MG: Why $R_3$? Probably $R3$ is better?}
Fig.~\ref{fig:compare-losses} illustrates how this Eq.~\ref{eqn:loss-coral} is used to compute the proposed CORAL loss function. % differ from each other\enter{, in terms of information processing steps??} \bs{This part seems a bit incomplete.}

\subsection{Training Algorithm}
\label{sec:training-policy}

\begin{algorithm}
\scriptsize
\caption{Training Algorithm for Seq2Seq Models using CORAL Loss (Mix-policy)}\label{alg:coral}
\begin{algorithmic}
\State $D = \{(c,r^+)_i\}_{i=1}^n$ \Comment{$n$ Positive pairs from training dataset}
\State $\theta^{(0)} \gets \text{Initialize Seq2Seq network weights}$
% \State $R \gets $ Pool of all utterances in dataset (train split)
\For{$(c,r^+)\in D$} \Comment{Actual implementation uses batch gradient descent}
    \If{$rand() > p_+$}
        \State \#Nucleus Sampling
        \State $r\leftarrow$ Sample $r^- \sim $ Nucleus$(c|\theta^{(k)})$ %\Comment{Nucleus Sampling}
        \State \# [or] RandomNegative Sampling
        \State $r\leftarrow$ Sample $r^- \sim $ Uniform(Training Utterance Pool) %\Comment{RandomNegative Sampling}
    \Else
    \State $r\leftarrow$ $r^+$ \Comment{Use Positive Response}
    \EndIf
    \State $R_3(c,r)=f_R(c,r) - m$ \Comment{$f_R(c,r)$: Response Retrieval Model Score}
    \State \#Compute Decoder output token distribution 
    \State Compute $P(r_t|r_{<t},c) \forall t \in [1,T]$ %\Comment{$P_{S2S}$: Decoder output token distribution}
    \State $L_{\text{CORAL}}=-R_3(c,r)\sum_{t=1}^{T}\log P(r_t|r_{<t},c)$
    \State \#Update parameters of $P$ using gradient descent on $L_{\text{CORAL}}$
    \State $\theta^{(k+1)} \gets \theta^{(k)} - \eta \nabla_{\theta^{(k)}} L_{\text{CORAL}} $
\EndFor
\end{algorithmic}
\end{algorithm}

\paragraph{Mix-Policy Training Using CORAL Loss}

RL training can be either on-policy or off-policy, depending on whether samples are 
generated from the parameterized policy network (the Seq2Seq model in our case) or 
obtained from a dataset of human generated examples.
% To do on-policy training using the proposed learning framework, we tried the 
% following steps. 
For pure on-policy training, we will have to rely on response sequences randomly sampled 
from the decoder. But, because of the combinatorial complexity of the response space% 
% (that could be generated by the decoder)
, it is highly unlikely that we would obtain 
any valid utterances/response candidates during on-policy training. 
Off-policy learning, on the other hand, becomes reliant on the ground truth responses only and cannot leverage any benefits of exploration. 
A mix of on-policy and off-policy training (henceforth referred as mix-policy) can potentially leverage both on-policy samples for better exploration and the ground truth responses for stabilizing the RL training process. Thus, to direct 
the model towards generating grammatically and semantically valid utterances, we perform mix-policy sample generation. To control the amount of mixing, we 
introduce a hyperparameter called $p^+$ (detailed below). The contributions from these two different types of samples are mixed as per the following equation. We also weigh positive response samples from the dataset using $R_3(c,r)$.
\begin{align}
    L_{\text{CORAL}}=-R_3(c,r^-) \sum_{t=1}^{T^-} \log P(r^-_t|r^-_{<t},c) -R_3(c,r^+) \sum_{t=1}^{T^+} \log P(r^+_t|r^+_{<t},c)
    \label{eqn:loss-coral-reweigh}
\end{align}
where $r^-$ and $r^+$ are response candidates generated by the policy (which is referred as on-policy) and from the ground truth dataset (which is referred as off-policy) respectively, and $T^-$ and $T^+$ are the lengths of the responses after tokenization.
Since random sampling based 
decoding can sometimes generate very low probability tokens while generating a sequence, we use 
nucleus sampling~\cite{holtzman2019curious} instead of random sampling for generating on-policy samples. 
We also tried a method to exclusively utilize more off-policy samples, which we refer to as Random Negative sampling (or RNS), for generating more diverse response samples during training. 
In RNS, we randomly sample utterances from 
the pool of all training-set utterances and use it as a response candidate in the training process.
In Algorithm \ref{alg:coral}, we show the steps used for training a Seq2Seq model 
using the CORAL loss on a dialog dataset $D$. 
% Fig.~\ref{fig:s2s-coral} \pg{No more, please rewrite as needed} 
% The right part of Fig.~\ref{fig:compare-losses} shows the architecture of our Seq2Seq model, which 
% generates a response given the context using a Transformer encoder-decoder model trained under 
% CORAL loss. 
We will refer to seq2seq models trained using the CORAL-loss as CORAL models.
% We illustrate the algorithm to train our Seq2Seq dialog model using minibatch gradient descent with 
% CORAL loss in Algorithm \ref{alg:coral}.

% \subsection{Generating on-policy samples}

\noindent\textbf{Hyperparameters of CORAL:}\label{hyperparameters-of-coral}
%CORAL comes with quite a few hyperparameters that 
%One needs to tune the following hyperparameters when using CORAL for training.
% and may depend on the retrieval model.
%\begin{itemize}
\textbf{(1) Probability of positive samples ($p_+$)} denotes the probability
    with which we use the ground truth (off-policy) response samples during mix-policy training.
    % During training 
    % each context in a batch is paired up with a response candidate from the full 
    % utterance pool (from training set). With probability $p_+$, we select the 
    % true response, and with probability $(1-p_+)$, we randomly select a utterance as response.
\textbf{(2) Margin ($m$)} denotes the minimum reward that we expect from model generations. 
    We use a fixed margin\footnote{More flexible versions of margin are possible based 
    on a learned value function (baseline function) or a 
    critic function (in Actor-Critic formulations). We 
    leave integration of such more advanced RL algorithms with the proposed CORAL-based learning framework as a future research direction.} value as the baseline reward for the RL training.  

 Although CORAL is derived from quite a different viewpoint, under certain hyperparameter settings ($m=0, p^+=1, R_3 \in (0,1)$) CORAL approximates a sample-weighted version of the CE loss. Also, training of a dialog generation model using CE may over-weigh generic responses more than the informative ones. A more detailed account of the similarities/differences between CORAL and CE are in the appendix. %In summary, CORAL incorporates quality of the response as measured by a response retrieval model as a factor in computing the loss.

\section{Experimental Setup}\label{sec:experimental-setup}

\subsection{Model Setup}\label{sec:model-architecture}

We use a standard transformer-based Seq2Seq (S2S) architecture for training\footnote{Libraries used: PyTorch, transformers, wandb, pytorch-lightning} a dialog generation model using the CORAL loss as shown in right part of Fig.~\ref{fig:compare-losses}. %We use PyTorch's inbuilt
% Transformer module to implement the model and did not change anything
% else in the architecture. 
% For tokenization and vocabulary, 
We train our models using early-stopping, up to a maximum of 50 epochs, based on validation-$R_3$ (average $R_3$ score of generated responses on the validation set). We use Adam optimizer with a peak learning rate of $10^{-4}$ that is warmed up (first 1000 steps) and decayed linearly. We use a single NVIDIA V100-32GB GPU-based systems for training our individual CORAL models.

\noindent\textbf{Retrieval Models} 
ESIM has two LSTM-based encoder layers for encoding the context and a candidate response, interleaved by a cross-attention layer. The sigmoid output from the ESIM model is used as the score for a context-response pair. 
Since BERT is a pretrained transformer-encoder model, we add a one-hidden layer MLP on top of the [CLS] token embedding. These models are finetuned on corresponding dialog datasets for the response retrieval task.
%Since both BERT and DMI are pretrained transformer-encoder models, we add an MLP (two hidden layers, with the final scalar output going through a sigmoid activation function) on top of the [CLS] token embedding. These models are finetuned on corresponding dialog datasets for the response retrieval task.
We obtain the reward ($R_3$) by subtracting the margin ($m$) (between $0$ and $1$) from the output score (sigmoid output). 

% \noindent\textbf{ESIM} consists of encoding and composition layers comprising single layer LSTMs for encoding the context and the response with a cross-attention layer in between.
% \noindent\textbf{BERT and DMI} We use the ``bert-base-uncased'' and the ``DMI\_Medium'' checkpoints, respectively. 
\noindent\textbf{CORAL} We implemented the dialog generation model using transformer-based Seq2Seq models with 6 self-attention layers, 8 self-attention heads and 1,024 as the size of hidden representations, for both the encoder and the decoder. We use the wordpiece tokenizer from BERT.

\noindent\textbf{CORAL-BB} This variant of the CORAL model is initialized with the pretrained-weights from the \textit{facebook/Blenderbot-400M-distill} checkpoint which allows us to leverage the power of the large-scale pre-training while also fine-tuning it using the proposed RL-based training algorithm. CORAL-BB uses the same architecture and tokenizer as the Blenderbot model (2 encoder and 12 decoder layers).

% \noindent \textbf{(4) Initialization:} We experiment with random initialization (CORAL) as well as initialization using pretrained checkpoint of \textit{facebook/Blenderbot-400M-distill} (CORAL-BB)

% \noindent\textbf{Training Setup}
%  We used early stopping based on a validation set for both R3 and S2S
% training.
%\end{itemize}

% \usepackage{booktabs}
% Train Len 76052
% Valid Len 7069
% Test Len 5740
% Train Len 470860
% Valid Len 23478
% Test Len 3247

\subsection{Datasets}\label{sec:data-and-code}

We use DailyDialog (DD) \citep{li2017dailydialog} and DSTC7-Ubuntu \citep{yoshino2019dialog} datasets for all our experiments. DD is an open-domain dialog dataset in English. DSTC7-Ubuntu is a domain-specific dataset based on
conversations from a Linux IRC channel. We create context-response (CR) pairs by unrolling the dialogs while ensuring a minimum of two previous utterances in the context.
% and current utterance is used as the response.
DD contains 76052, 7069 and 5740 CR pairs for train, validation and test, resp., and DTSC7-Ubuntu contains 470860, 23478 and 3247 pairs, resp. However, \citet{wen-etal-2022-ddcc} recently showed that there are some leaks between the splits of the DD dataset, and proposed a cleaned version (DD$_c$) containing new splits of 60243, 6644, 5986 CR pairs for train, validation and test, resp.

\subsection{Baselines}\label{baselines}
% \pg{Please make a systematic division -- small scale (no pretraining) vs. pretrained models Use noindent textbf rather than itemize to save space.}
% \begin{itemize}
  
% \item
  
%   \textbf{S2S-CE} Transformer-based Seq2Seq \citep{vaswani2017attention} model trained for dialog
%   generation model using the CE loss. This uses exactly the same architecture as S2S-CORAL for fair comparison.
  
%   \item
\noindent\textbf{Non-pretrained Baselines:}
% \noindent 
(1) \textbf{Mirror}~\citep{li2021improving}: Seq2Seq model that extends CVAE~\citep{shen2017conditional}, and is trained with a backward-reasoning loss function. It optimizes for generating final and pre-final utterances in a bidirectional fashion. This
  is state-of-the-art loss function for training small-scale dialog generation models outperforming \citep{shen2017conditional,zhao2017learning,saleh2020hierarchical}.
%For making the model generate more informative responses, authors of Mirror extended the CVAE model for dialog generation ,zhao2017learning} to optimize for generating final and pre-final utterances in a bidirectional fashion. 
% \noindent
  (2) \textbf{AdaLabel}~\cite{wang2021diversifying}: Uses adaptive label smoothing and soft-target distribution to prevent the model from being overconfident over a single choice. 
  % It also uses a soft-target distribution depending on the context, instead of usual one-hot distribution.
% \item
% \noindent
  (3) \textbf{ALDGen}~\cite{li2017adversarial}: Includes a discriminator to differentiate between human-generated and machine-generated dialogs, and a generator to optimize the score given by the discriminator, using RL.

\noindent\textbf{Pretrained Baselines:} We consider zero-shot as well as fine-tuned variants of the following.
% \noindent
   (1) \textbf{Blenderbot}~\citep{roller2020recipes}: Transformer-based S2S model pretrained on a large dialog corpus
  based on Reddit and finetuned on Blended-Skill-Talk dataset \citep{smith2020together}. %This final finetuning helps to improve the quality of responses by reducing unwanted response generation.
% \item
% \noindent
  (2) \textbf{DialoGPT}~\citep{zhang2019dialogpt}: GPT-2 \citep{radford2019language} based language model further finetuned on dialogs from Reddit.
% \item 
% \noindent
    (3) \textbf{DialogRPT}~\citep{gao2020dialogrpt}: A response ranking model trained on a dataset of upvote/downvote and number of replies on Reddit comments. For generation, it reranks sampled responses from DialoGPT and returns the one with highest rank.
% \end{itemize}

\subsection{Evaluation Metrics}\label{evaluation-metrics}

We use a standard set of referenced evaluation metrics (BLEU, METEOR) and a few recently proposed reference-free metrics (MaUde, DEB) for automatic evaluation. 
BLEU \citep{papineni2002bleu} and METEOR \citep{banerjee2005meteor} measure lexical-overlap between n-grams of the predicted and ground truth response.
% however, they cannot always capture the validity of a generation.
MaUde \citep{Sinha2020LearningAU, yeh2021comprehensive} captures suitability between a context and a response without a ground truth reference. It is based on a response retrieval model and can be optimized through CORAL loss function. We report results for two MaUde variants based on ESIM \citep{chen2016enhanced,chen2019sequential} and BERT \citep{devlin2018bert} finetuned on the target datasets. DEB~\cite{Sai2020ImprovingDE} is a BERT-based dialog evaluation metric that is trained on a large-scale dataset of Reddit conversations (DEB(r)) and finetuned on a dataset of multiple relevant and adversarial responses for each context (DEB(a)). Distinct-n \citep{liu2016not} measures n-gram diversity in generated responses.

\section{Results and Discussions}\label{sec:results}

\begin{table*}[]
\caption{Results for DailyDialog (DD$_c$) and DSTC7-Ubuntu datasets: We can see that by optimizing the contextual $R_3$ score directly, using REINFORCE, the CORAL model is able to produce contextually relevant responses (as evident from MaUde and DEB). 
% Average token length of generated responses has also been reported.
% to check that the model is not resorting to short utterances, such as ``I don't think I know about \textit{[topic\_word]}'', just to be coherent.
%Among the baselines, Blenderbot-3B, which is a large-scale pretrained model, produces the most human-like responses as per MaUde but not as diverse as the CORAL-based models. 
$\text{CORAL}_x$ denotes a Seq2Seq model trained with CORAL loss. `$x$' identifies the retrieval model used for the $R_3$ reward. PT=Pretrained. %Note that we used the DialoGPT-medium, DialogRPT-medium, Blenderbot checkpoints. \textit{We have kept these large-scale baselines as a separate group, in the table.} Each value reported for CORAL models is computed as average of 5 runs.
} 
\label{tab:results-dd-new2}
\centering
\scriptsize
% \resizebox{\textwidth}{!}{%
\begin{tabular}{@{}p{0.3in}lclrrrrrrrrr@{}}
    \hline
        \multicolumn{13}{c}{\textbf{DailyDialog (DD$_c$)}}\\ \hline
            &Model & PT?&Size& Len & BLEU & METEOR & Dist-1 & Dist-2 & MaUde$_\text{ESIM}$ & MaUde$_\text{BERT}$ & DEB(r) & DEB(a) \\ \hline
    &Ground Truth &&& 11.7 & 0.997 & 0.986 & 0.067 & 0.410 & 0.733 & 0.828 & 0.898 & 0.933 \\ \hline
        \multirow{3}{0.3in}{(A) No pre-training}&Mirror &\xmark& 240M&3.8 & 0.018 & 0.026 & 0.016 & 0.059 & 0.535 & 0.505 & 0.516 & 0.740 \\
    &Adalabel&\xmark &90M& 11.6 & 0.069 & 0.052 & 0.031 & 0.189 & 0.589 & 0.459 & 0.627 & 0.720 \\
    &ALDgen&\xmark & 68M&10.6 & 0.054 & 0.048 & 0.024 & 0.184 & 0.542 & 0.326 & 0.528 & 0.653 \\ \hline
    \multirow{3}{0.3in}{(B) Zero-\\ shot}&DialoGPT (ZS)&\cmark& 345M & 8.7 & 0.068 & 0.047 & 0.042 & 0.169 & 0.693 & 0.630 & 0.842 & 0.896 \\
    &DialogRPT (ZS)&\cmark & 345M& 15.8 & 0.082 & 0.055 & 0.027 & 0.116 & 0.715 & 0.642 & 0.850 & 0.874 \\
    &Blenderbot (ZS)&\cmark& 365M & 17.1 & 0.104 & 0.077 & 0.025 & 0.115 & 0.697 & 0.671 & 0.958 & 0.968 \\ \hline
    \multirow{3}{0.3in}{(C) Fine-\\ tuned}&DialoGPT (FT)&\cmark& 345M & 6.1 & 0.070 & 0.064 & \textbf{0.060} & 0.265 & 0.775 & 0.799 & 0.908 & 0.953 \\
    &DialogRPT (FT)&\cmark& 345M & 17.3 & 0.101 & 0.089 & 0.029 & 0.148 & 0.839 & 0.839 & 0.951 & 0.962 \\
    &Blenderbot (FT)&\cmark& 365M & 29.8 & 0.117 & 0.110 & 0.027 & 0.147 & 0.874 & 0.831 & 0.962 & 0.957 \\ \hline
   \multirow{8}{0.3in}{(D) Our \\Models} &CORAL$_\text{ESIM}$ (offp)&\xmark& 93M &10.2&0.091&0.062&0.043&0.271&0.816&0.591&0.823&0.840\\
&CORAL$_\text{ESIM}$ (mixp)&\xmark& 93M &4.8&0.047&0.040&0.028&0.140&\textbf{0.918}&0.537&0.857&0.892\\
&CORAL$_\text{BERT}$ (offp)&\xmark& 93M & 9.2 & 0.088 & 0.060 & 0.045 & \textbf{0.274} & 0.679 & 0.595 & 0.763 & 0.827 \\
    &CORAL$_\text{BERT}$ (mixp)&\xmark& 93M & 6.0 & 0.061 & 0.045 & 0.033 & 0.184 & 0.704 & 0.650 & 0.775 & 0.879 \\ 
 &CORAL-BB$_\text{ESIM}$ (offp)&\cmark& 365M &21.3&0.126&0.110&0.034&0.178&0.896&0.874&0.984&0.984\\
&CORAL-BB$_\text{ESIM}$ (mixp)&\cmark& 365M&20.8&0.124&0.108&0.031&0.171&0.875&0.852&0.984&\textbf{0.986}\\
    &CORAL-BB$_\text{BERT}$ (offp)&\cmark& 365M & 21.2 & 0.125 & \textbf{0.112} & 0.035 & 0.186 & 0.856 & 0.881 & 0.978 & 0.978 \\
    &CORAL-BB$_\text{BERT}$ (mixp)&\cmark& 365M & 21.1 & \textbf{0.128} & 0.111 & 0.032 & 0.170 & 0.878 & \textbf{0.888} & \textbf{0.986} & \textbf{0.986} \\
    \hline \hline
    \multicolumn{13}{c}{\textbf{DSTC7-Ubuntu}}\\ \hline
    &Ground Truth &&& 13.7 &  0.974&0.967  & 0.092 & 0.523 & 0.858 & 0.736 & 0.906 & 0.929 \\ \hline
    \multirow{3}{0.3in}{(A) No pre-\\ training}&Mirror &\xmark& 240M& 6.2 & 0.030 & 0.039 & 0.007 & 0.015 & 0.512 & 0.493 & 0.682 & 0.853 \\
    &Adalabel&\xmark & 90M&15.7 & \textbf{0.202} & \textbf{0.182} & 0.054 & 0.310 & 0.793 & 0.705 & 0.877 & 0.915 \\
    &ALDgen &\xmark& 68M&11.6 & 0.058 & 0.045 & 0.076 & \textbf{0.442} & 0.448 & 0.487 & 0.709 & 0.782 \\ \hline
    \multirow{3}{0.3in}{(B) Zero-shot}&DialoGPT (ZS)&\cmark & 345M&8.6 & 0.056 & 0.037 & 0.050 & 0.172 & 0.706 & 0.656 & 0.839 & 0.901 \\
    &DialogRPT (ZS)&\cmark & 345M&14.2 & 0.071 & 0.042 & 0.035 & 0.133 & 0.733 & 0.681 & 0.845 & 0.883 \\
    &Blenderbot (ZS)&\cmark& 365M&17.8 & 0.087 & 0.057 & 0.027 & 0.089 & 0.568 & 0.529 & 0.778 & 0.746 \\ \hline
    \multirow{3}{0.3in}{(C) Fine-tuned}&DialoGPT (FT)&\cmark&345M&6.2&0.053&0.046&\textbf{0.110}&0.402&0.821&0.661&0.855&0.888\\
    &DialogRPT (FT)&\cmark&345M&13.6&0.090&0.063&0.071&0.322&0.873&0.749&0.933&0.939\\
    &Blenderbot (FT)&\cmark&365M&17.0&0.121&0.091&0.079&0.382&0.824&0.780&0.982&0.963\\
    % DialoGPT-new & 6.2 & 0.0527 & 0.046 & 0.1101 & 0.4017 & 0.8210 & 0.6610 & 0.855 & 0.888 \\
    % DialogRPT-new & 13.6 & 0.0902 & 0.063 & 0.0710 & 0.3219 & 0.8731 & 0.7485 & 0.933 & 0.939 \\
    % blenderbot-400M-distill & 17.0 & 0.1210 & 0.0905 & 0.0788 & 0.3821 & 0.8237 & 0.7800 & 0.982 & 0.963 \\ 
    \hline
    \multirow{8}{0.3in}{(D) Our \\Models}&CORAL$_\text{ESIM}$ (offp)&\xmark & 93M&10.3&0.173&0.158&0.078&0.425&0.859&0.760&0.930&0.949\\
    &CORAL$_\text{ESIM}$ (mixp)&\xmark & 93M&8.5&0.134&0.122&0.075&0.403&0.869&0.753&0.929&0.950\\
    &CORAL$_\text{BERT}$ (offp)&\xmark & 93M&10.5 & 0.095 & 0.072 & 0.073 & 0.412 & 0.853 & 0.811 & 0.955 & 0.958 \\
    &CORAL$_\text{BERT}$ (mixp)&\xmark & 93M&12.4 & 0.097 & 0.068 & 0.068 & 0.364 & 0.879 & \textbf{0.848} & 0.969 & 0.973 \\
    &CORAL-BB$_\text{ESIM}$ (offp)&\cmark& 365M&15.5&0.106&0.067&0.057&0.253&0.924&0.787&0.990&0.986\\
&CORAL-BB$_\text{ESIM}$ (mixp)&\cmark & 365M&14.5&0.106&0.066&0.056&0.232&\textbf{0.967}&0.810&\textbf{0.996}&0.989\\
    &CORAL-BB$_\text{BERT}$ (offp)&\cmark& 365M&15.3 & 0.104 & 0.066 & 0.054 & 0.241 & 0.896 & 0.793 & 0.989 & 0.986 \\
    &CORAL-BB$_\text{BERT}$ (mixp)&\cmark & 365M&15.2 & 0.107 & 0.066 & 0.048 & 0.205 & 0.923 & 0.831 & 0.995 & \textbf{0.991} \\
 \hline
\end{tabular}
\end{table*}

\subsection{Automatic Evaluation}
% \textbf{Justification for why we are going to focus on MaUde and DEB:}
In Table~\ref{tab:results-dd-new2}, we present the automatic evaluation results for response generation using non-pretrained, zero-shot and finetuned baselines and our proposed models separately for both the DD$_c$ and DSTC7 datasets\footnote{We also report results on the original DD dataset in the Appendix.}. For our proposed models, we present variants based on (1) reward function (BERT/ESIM), (2) sample generation method (off-policy and mix-policy), (3) Random or Blenderbot initialization. 
For our proposed CORAL models, %we first trained the model on the DD and DSTC7 datasets and saved the best checkpoint based on the average $R_3$ reward on the validation set. Further, 
we found $m$=0.4 and $p^+$=0.8 for DSTC7-Ubuntu and $m$=0 and $p^+$=0.8 for DD$_c$ as the best hyperparameters. For CORAL-BB, we found the best values to be $p^+$=0.6, $m$=0.2. Fig. 1 in Appendix shows sensitivity analysis for $p_+$ and $m$. % for both the datasets.
% HARDCODED

The focus of this work has been to propose a framework that can train dialog generation models by optimizing estimates of human preference. Prior works \cite{liu2016not,yeh2021comprehensive} have shown that reference-based metrics (e.g., BLEU, METEOR) have poor correlation with human ratings. Hence, we focus on evaluating models against reference-free relevance metrics such as MaUde and DEB. 

In terms of MaUde and DEB, models of the CORAL family outperform all baselines by a significant margin. The diverse and high quality responses justify the choice and design of the CORAL loss function. Especially our Blenderbot initialized CORAL-BB models (in block D of Table~\ref{tab:results-dd-new2}) outperform all pretrained dialog generation models that use standard CE loss based finetuning. Because of the extensive pretraining done, we do not expect non-pretrained CORAL models to fully outperform pretrained baselines. Still, we observe that CORAL is able to beat all the zero-shot models for DSTC7-Ubuntu. Further, for DSTC7-Ubuntu, surprisingly, our non-pretrained CORAL models outperform even strong finetuned baselines which are $\sim$4x in size. These results indicate that, by using an estimate of human preference as the training objective, it is possible to train models that are specifically good at achieving better relevance scores.

% On the other hand, although CORAL models are much smaller in size than pretrained baseline models, they still outperform the zero-shot pretrained models in the relevance measures. 
%  \\

Amongst variants of the CORAL family, we observe that mix-policy training often achieves higher relevance scores than off-policy training, indicating the benefits of allowing exploration. The off-policy training approach, which essentially reweighs response samples in CE loss using the R$_3$ reward, also performs better than the baseline models trained using standard CE loss. This shows that not all samples from the training dataset contribute equally or positively to CE-based training. Simply downsampling certain samples using the R$_3$ reward helps us train dialog generation models that can generate responses more relevant to the context.

%In our work, we also train a set of CORAL models in which we initialize the parameters of the transformer encoder and decoder using a distilled checkpoint of the large-scale Blenderbot-3B model. Following this initialization, we finetune the Seq2Seq model using the CORAL loss function on any of the three datasets. 
% This approach allows us to leverage the power of the large-scale pre-trained model while also fine-tuning it using the proposed RL-based training algorithm. (Best hyperparameters: $Mix-policy, p^+=0.6, m=0.2$)} \pg{BB-3B or 400M?}

Diversity of the generated responses is indicated by the Dist-1 and Dist-2 metrics. We observe that non-pretrained CORAL models have better diversity compared to pretrained ones. We also observe that, for CORAL-BB models, output length and diversity are inversely related. This could be because as the model outputs longer generations, it tends to reuse similar ngrams.

% All baselines, except AdaLabel, have very low diversity for the generated responses. The LSTM-based Mirror model mostly generates the same bland response (e.g., \textit{``I don't know how to do that''}, \textit{``I don't know what to suggest''}) for most contexts leading to a low diversity score. 

%Next, we also report the \textbf{BLEU} and \textbf{METEOR} scores which are two standard metrics used for evaluating NLG models. Though the reader should note that these 

Our results in Table \ref{tab:results-dd-new2} also demonstrate the disconnect between lexical-overlap metrics (BLEU, METEOR) and the relevance metrics (MaUde and DEB) as shown in several prior works \citep{liu2016not,Sinha2020LearningAU,yeh2021comprehensive}. For DSTC7-Ubuntu, Adalabel achieves highest BLEU and METEOR scores despite the poor relevance values. On the other hand, for DD$_c$, the CORAL model with the highest MaUde also attains the highest BLEU score. 

In terms of avg. length of the generated responses, the general trend is that the pretrained models outperform other baselines (i.e., trained from scratch) and the same holds true for CORAL models. Though it should be noted that lengthier generation does not necessarily mean better quality responses.

We trained two types of baseline as well as CORAL models. One set of models have been initialized (for finetuning or used as is for zero-shot) from pretrained dialog generation models. We observe that pretraining gives a significance boost to the relevance scores of the model among the CE-based baselines (block A versus B in Table~\ref{tab:results-dd-new2}). The second set of improvements in relevance score comes from finetuning these pretrained models on the task dataset. This can be observed by comparing the zero-shot pretrained models with the fine-tuned ones (block B versus C in Table~\ref{tab:results-dd-new2}). And the final improvements are obtained when these pretrained models are finetuned using the CORAL loss function, which can be observed by comparing Blenderbot-finetuned versus CORAL-BB. 

% Large-scale (in terms of data or model parameter size) pretrained language models have become a very prominent candidate for open-domain dialog generation models with the development of models like DialoGPT, Blenderbot, etc.

% Thus, we also compare our CORAL models with two variants of these large-scale models: zero-shot and finetuned versions. 

% Even though DD and Reddit-based pretraining datasets both comprise open-domain English conversations, $\text{CORAL}_\text{BERT}$ outperforms the zero-shot variants on DD$_c$ and DSTC7-Ubuntu. 

% $\text{CORAL}_\text{BERT}$ is even able to beat the finetuned version of DialoGPT and Blenderbot on the DSTC7-Ubuntu dataset. This proves the strength of the loss function and RL training paradigm proposed in this paper for dialog generation models.

\subsection{Hyperparameter Sensitivity Analysis}
To better understand the effects of various hyperparameters (see Section \ref{sec:training-policy}) on the final trained model, we perform extensive experiments by varying $p^+$, $m$ and sampling method. 
% HARDCODED
The complete set of results is displayed in Fig. 1 in the appendix. All the comparisons are done based on the best average reward obtained by the model on validation set. In general, the mix-policy setup outperforms off-policy training routines. For DD, lower margin values tend to have higher $R_3$ scores. But, for DSTC7-Ubuntu and DD$_c$, in case of mix-policy training, the $R_3$ increases with margin. We observed that mix-policy training using nucleus sampling performed better than with RandomNegatives. 
%Off-policy training worked better with positive ground truth responses only, with RandomNegatives generally having a detrimental effect on the final reward achieved by the model.

\subsection{Human Evaluation Study}
\label{sec:human-eval}

\begin{figure}[!htbp]
\centering
\includegraphics[width=0.45\linewidth]{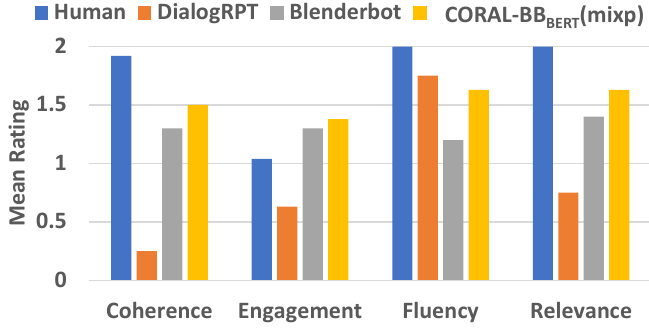}
  \caption{Human evaluation results: Avg. ratings for Coherent, Engagement, Relevance, and Fluency.}
  \label{fig:human-eval}
\end{figure}

As automatic evaluation metrics cannot capture all the nuances of how humans assess a model generated response, we also run a crowdsourced human evaluation study for various models. We used the Appen.com platform to run these surveys. Three different annotators based out of USA, rated a context-response pair in terms of engagement, relevance, fluency and coherence on a 0-2 scale: No (0), Somewhat (1), Yes (2). Detailed annotation questionnaire is in the Appendix. This evaluation process was run on 15 randomly selected contexts from DD$_c$ test set, for Blenderbot, DialogRPT and CORAL-BB$_\text{BERT}$ (mixp) models. %For each context and a model generated response, we obtained judgments from three different annotators.

We plot mean ratings from the annotators, for each of the aspects for each model in Figure \ref{fig:human-eval}. Overall, it can be seen that the CORAL-BB model has higher scores compared to DialogRPT and Blenderbot on coherence and relevance, but lower than human generated dialog. DialogRPT has the highest fluency score indicating that it produces most well formed/grammatical English sentences, with CORAL in the second position with comparable scores. One peculiar finding was that Blenderbot as well as CORAL-BB are more engaging than groundtruth as can be also seen in examples shown in Table~\ref{tab:manual-samples} (discussed further in Section \ref{sec:case-study}). 

%shows that the CORAL-based models outperform the baselines in all the three aspects. CORAL-mixp (nucleus) outperforms CORAL-offp in engagement and relevance, whereas CORAL-offp is more fluent than CORAL-mixp.

% The human generated dialog had the best scores overall, but the CORAL model still performed well on all the metrics, showing that it is capable of generating coherent, engaging, fluent, and relevant dialog.

\begin{table*}[]
  % \caption{Sample generations from baselines (DialogRPT (FT), Blenderbot (FT)) and $\text{CORAL-BB}_\text{BERT}$(mixp).}
  \caption{Sample generations from DialogRPT (FT), Blenderbot (FT) and $\text{CORAL-BB}_\text{BERT}$(mixp).}
  % The individual sentences within an utterance are not very consistent with each other.
  \label{tab:manual-samples}
  \centering
  \scriptsize
%   \resizebox{\textwidth}{!}{%
  \begin{tabular}{p{0.14\textwidth}|p{0.26\textwidth}|p{0.31\textwidth}|p{0.22\textwidth}}
  \hline
  \multicolumn{4}{p{5.4in}}{\textbf{Context}: What exactly is wrong with it ? EOU The phone turns off by itself . EOU That's fine , but do you have your receipt with you ? EOU} \\ \hline
  \textbf{Human}: Yes , here it is . & 
  \textbf{DialogRPT}: Yes, I do. I'll just take it to the Apple Store. Thanks. I'll be back in a minute. Bye. & 
  \textbf{Blenderbot}: I'm afraid I don't have it with me. Can you give me your name, phone number, and phone number so I can make sure that I can get it back to me in the next day or two? If not, I won't be able to return it to you. & 
  \textbf{CORAL-BB$_\text{BERT}$ (mixp)}: Yes, I have my receipte with me. I'll return it to you in a few minutes. \\ \hline

  \multicolumn{4}{p{5.4in}}{\textbf{Context}: I like to play computer games , read books , go shopping , and play pool . EOU Out of those what is your favorite ? EOU} \\ \hline
  \textbf{Human}: My favorite is to play computer games .  & 
  \textbf{DialogRPT}: I like to play computer games, read books, go shopping, and play pool. & 
  \textbf{Blenderbot}: I don't have a favorite. I like all of them. What do you think of the new Star Wars movie? I think it's going to be great! & 
  \textbf{CORAL-BB$_\text{BERT}$ (mixp)}: I like playing video games, too. I also like going to the theater to see a movie. \\ \hline

  \multicolumn{4}{p{5.4in}}{\textbf{Context}: Come on in . Let me show you around . Here ’ s the living room .  EOU  Great . The view from this window is wonderful .  EOU} \\ \hline
  \textbf{Human}: This is the kitchen . & 
  \textbf{DialogRPT}: Thank you. I ’ ll take a look. I ’ ll be back in a few minutes. & 
  \textbf{Blenderbot}: Thank you very much. I ’ ll be back in a few minutes. Would you like anything to drink? & 
  \textbf{CORAL-BB$_\text{BERT}$ (mixp)}: I ’ m glad you like it. It ’ ll be my first time living in a new apartment. \\ \hline

  \multicolumn{4}{p{5.4in}}{\textbf{Context}: Are you ready for the camping trip ?  EOU  All set . I've got my makeup , my Cds , and my portable TV set ...   EOU} \\ \hline
  \textbf{Human}: Hello ! We're supposed to be getting back to nature .& 
  \textbf{DialogRPT}: You're going to have a great time! I'm sure you'll have a great time! I'm sure you'll have a great time! & 
  \textbf{Blenderbot}: Camping trips are so much fun. You're going to have a great time!  & 
  \textbf{CORAL-BB$_\text{BERT}$ (mixp)}: What are you going to do in the woods?  \\ \hline
  
  \end{tabular}
\end{table*}

\subsection{Case Study: Generation Quality}
\label{sec:case-study}
% \textcolor{red}{MG: there is no manual evaluation here. we should call this section as case studies.}
%We also manually went through a random selections of generated outputs from all models and compared them. We wanted to find out the strengths and weaknesses of each of the approaches, both for baselines and CORAL. We summarize these points below. 
We provide samples generated from $\text{CORAL-BB}_\text{BERT}$(mixp) and some baseline models (DialogRPT (FT), Blenderbot (FT)) in Table \ref{tab:manual-samples}. We find that responses generated by CORAL present information more consistently (coherence) than the baselines (example 1). We also notice that DialogRPT is more repetitive and in some cases just repeats a previous utterance (example 4). The examples also show that Blenderbot and CORAL-BB are more engaging then DialogRPT as well as ground truth responses. In general, we find that CORAL-BB responses are more conversational; the responses show that the speaker is interested in the other person's opinion and is willing to continue the conversation (examples 3 and 4).

\textbf{Error analysis}: To gain insights into the limitations of CORAL, we conducted an error analysis on responses generated by our best model across two datasets. Broad error buckets include difficulties in correctly attributing the next responses to user vs bot, tendency to generate relevant but divergent follow-ups, and occasional generation of consecutive utterances as a single response. More details are in Appendix.

\section{Conclusion}
\iffalse
Natural Language Generation models in NLP have usually been trained using the cross-entropy loss, and it has been reasonably
successful too. But dialog-response generation poses a few
unique challenges for the CE loss. 
First, cross-entropy loss assumes
that for any given input, the only possible output is the one available
as the ground truth through the dataset. 
Secondly, cross-entropy does not take the context into consideration
while processing the response. 
\fi

In this paper, we
proposed CORAL, a novel loss function to circumvent 
shortcomings of CE loss for dialog generation. Specifically, using mix-policy based training in CORAL, we can train dialog generation models without assuming the ground-truth as the only correct response,
% only one fixed ground-truth response 
%\pg{But we did not show on-policy, can we still claim this?} 
and the value of the loss function is based on both the context and the response. 
The CORAL loss is based on pretrained response retrieval models that, in prior literature, have been shown to correlate with human preferences. Experiments over two diverse benchmarks have shown that it comprehensively outperforms other strong baseline models (non-pretrained, zero-shot as well as finetuned). We plan to extend this framework for more efficient training using advanced RL methods and ideas from curriculum learning.
Finally, we hope that our work will open up interesting areas for future research about how to train better reward functions that can capture other aspects of response qualities.

% for designing a learning curriculum for RL-based training using the mix-policy method. This will help make the training of the dialog generation model more efficient.
%The proposed loss function will make it possible to train future models focused on maximizing human preference. 
% We hope that our work will motivate the NLP
% community to look for more suitable loss functions for training dialog generation models. 

% We plan to extend this framework for training larger scale models that can capture more patterns from larger training data.  

 %\textcolor{red}{MG: also use some distillation based setup?}

%Research in loss functions also gives us insights into which attributes for a generative dialog system are most appreciated in the final model as we compare their outcomes.

%\section*{Acknowledgments}
%This work was partially supported by Microsoft Academic Partnership Grant (MAPG) 2021. The first author was also supported by Prime Minister's Research Fellowship (PMRF), India.

% \bibliographystyle{abbrvnat}
\bibliography{main}

\newpage

\appendix

\section{Similarities and Differences between CORAL and CE Loss Functions}
In this section, we explore the similarities and differences between the proposed CORAL loss and the CE loss function. Although CORAL is derived from quite a different viewpoint, under certain hyperparameter settings CORAL approximates a weighted version of the CE loss.

\begin{enumerate}
    \item If we only consider positive samples as candidate responses and set the score range ($score\in[0,1]$) and margin m ($m=0$) such that $R_3$ is always greater than zero, CORAL is equivalent to a weighted version of CE.
    \item Cross-entropy loss has always relied \textit{strictly} on the positive responses in the dataset. CORAL utilizes both positive and negative response candidates. 
    \item
      Training of a dialog generation model using CE may over-weigh generic responses more than more informative ones as there is no mechanism for automatically assigning weights to different $\langle$context,response$\rangle$ pairs.
      CORAL has provision for assigning different weights for different $\langle$context, candidate response$\rangle$ pairs.
    \item
      CORAL uses randomly sampled response candidates for training which allows us to utilize more samples of $\langle$context,response$\rangle$ pairs during training.
      This provides a richer training signal from the same dataset.
    \item
    CE loss decomposes to a token level comparison between the predicted and the target token. Its main goal is to increase the probability of the tokens in ground truth response strictly in the given form and order.
    CORAL loss works quite differently as it treats responses as whole units. It will either increase or decrease probability of responses as a whole, based
    on their semantics and compatibility to the context.
\end{enumerate}

\section{Hyperparameter Sensitivity Analysis}
% To better understand the effects of various hyperparameters on the final trained model, we perform extensive experiments by varying $p^+$, $m$ and sampling method. 
% The complete set of results is displayed in Fig.~\ref{fig:hparams} in the appendix. All the comparisons are done based on the best average reward obtained by the model on validation set. In general, the mix-policy setup outperforms off-policy training routines. For DD, lower margin values tend to have higher $R_3$ scores. But, for DSTC7-Ubuntu and DD$_c$, in case of mix-policy training, the $R_3$ increases with margin. We observed that mix-policy training using nucleus sampling performed better than with RandomNegatives. 

\begin{figure}
     \centering
     \begin{subfigure}[b]{0.48\textwidth}
         \centering
         \includegraphics[width=\textwidth]{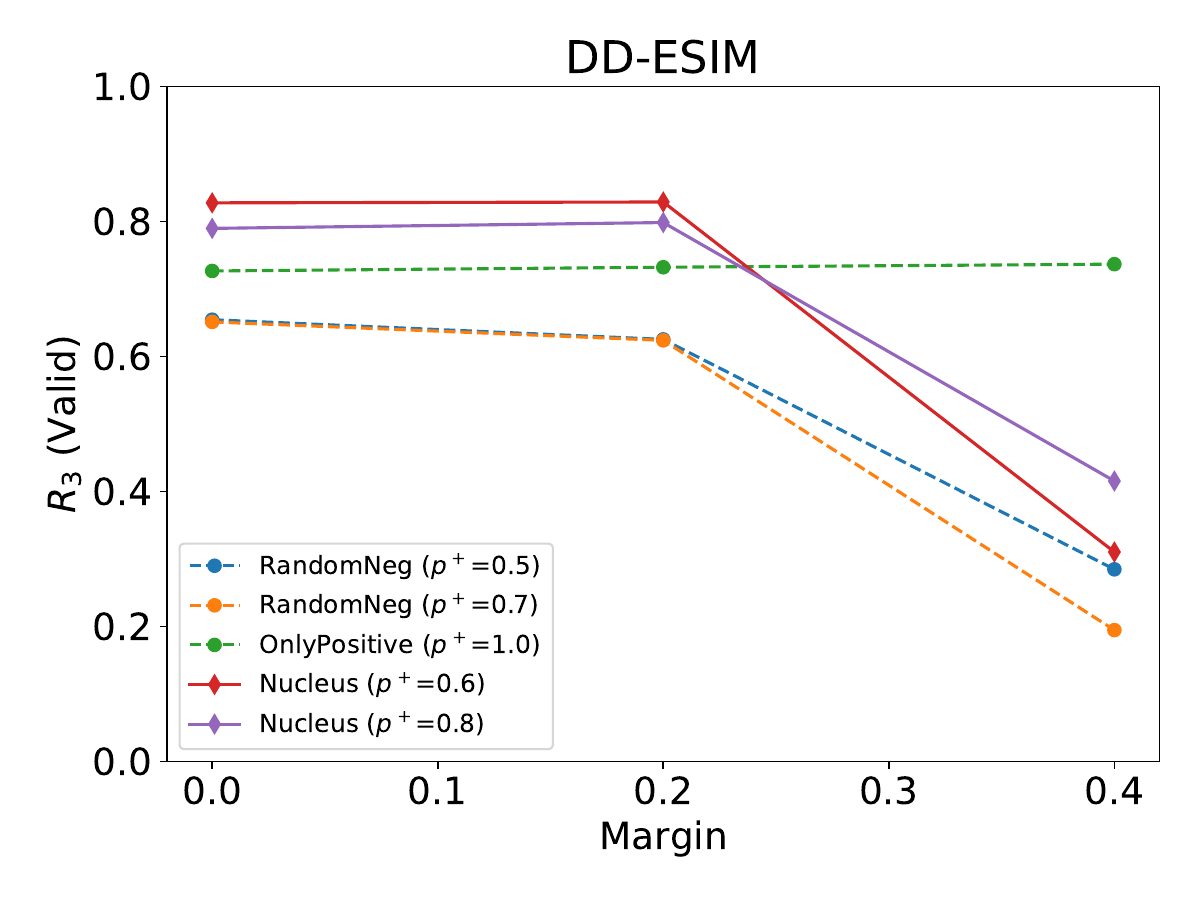}
         \caption{$\text{CORAL}_\text{ESIM}$ (DailyDialog)}
         \label{fig:dd1}
     \end{subfigure}
     \hfill
     \begin{subfigure}[b]{0.48\textwidth}
         \centering
         \includegraphics[width=\textwidth]{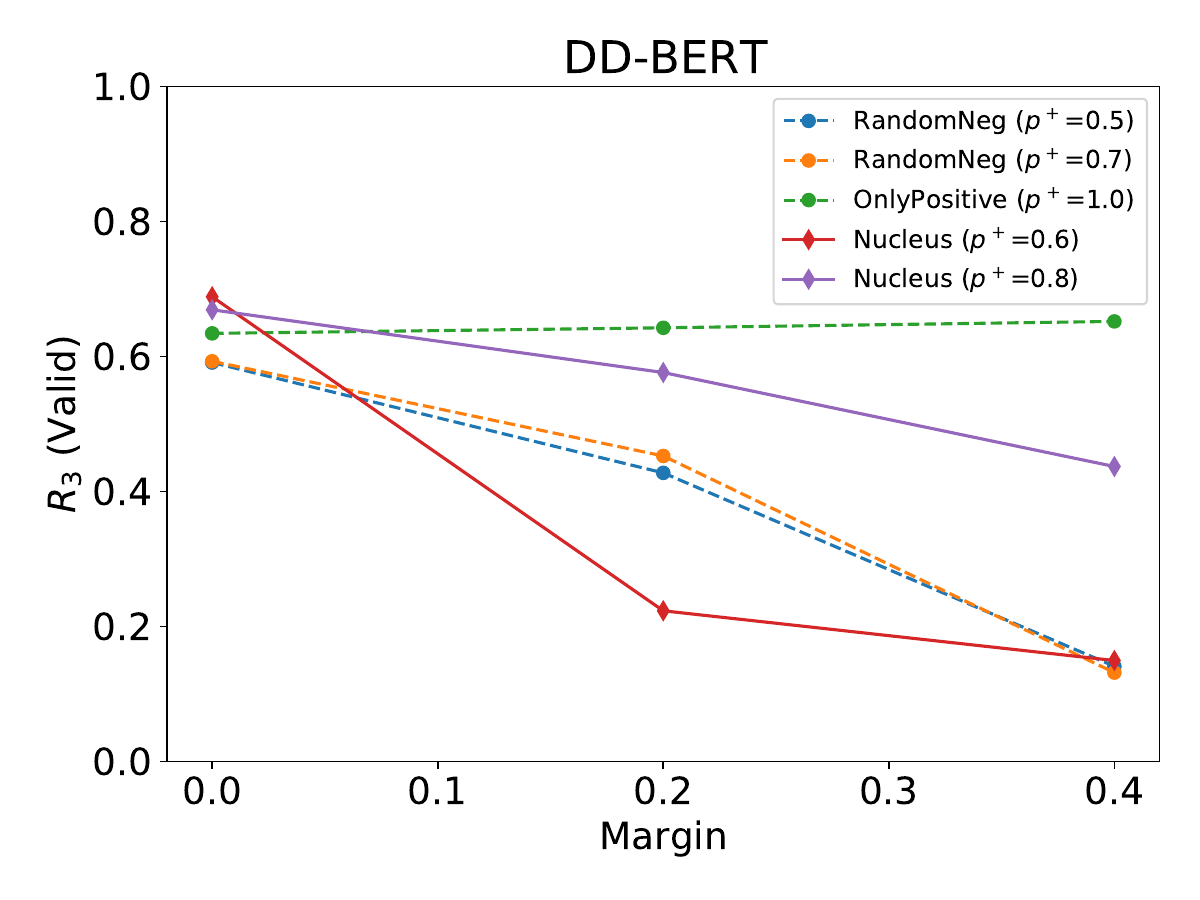}
         \caption{$\text{CORAL}_\text{BERT}$ (DailyDialog)}
         \label{fig:dd2}
     \end{subfigure}
     
     % \begin{subfigure}[b]{0.48\textwidth}
     %     \centering
     %     \includegraphics[width=\textwidth]{plots/dd-DMI-merged-hparams.pdf}
     %     \caption{$\text{CORAL}_\text{DMI}$ (DailyDialog)}
     %     \label{fig:dd3}
     % \end{subfigure}
     % \hfill
%     \caption{Hyperparameter Sensitivity Analysis (DailyDialog)}
%     \label{fig:dd-hparams}
% \end{figure}
% \begin{figure}
%      \centering
     \begin{subfigure}[b]{0.48\textwidth}
         \centering
         \includegraphics[width=\textwidth]{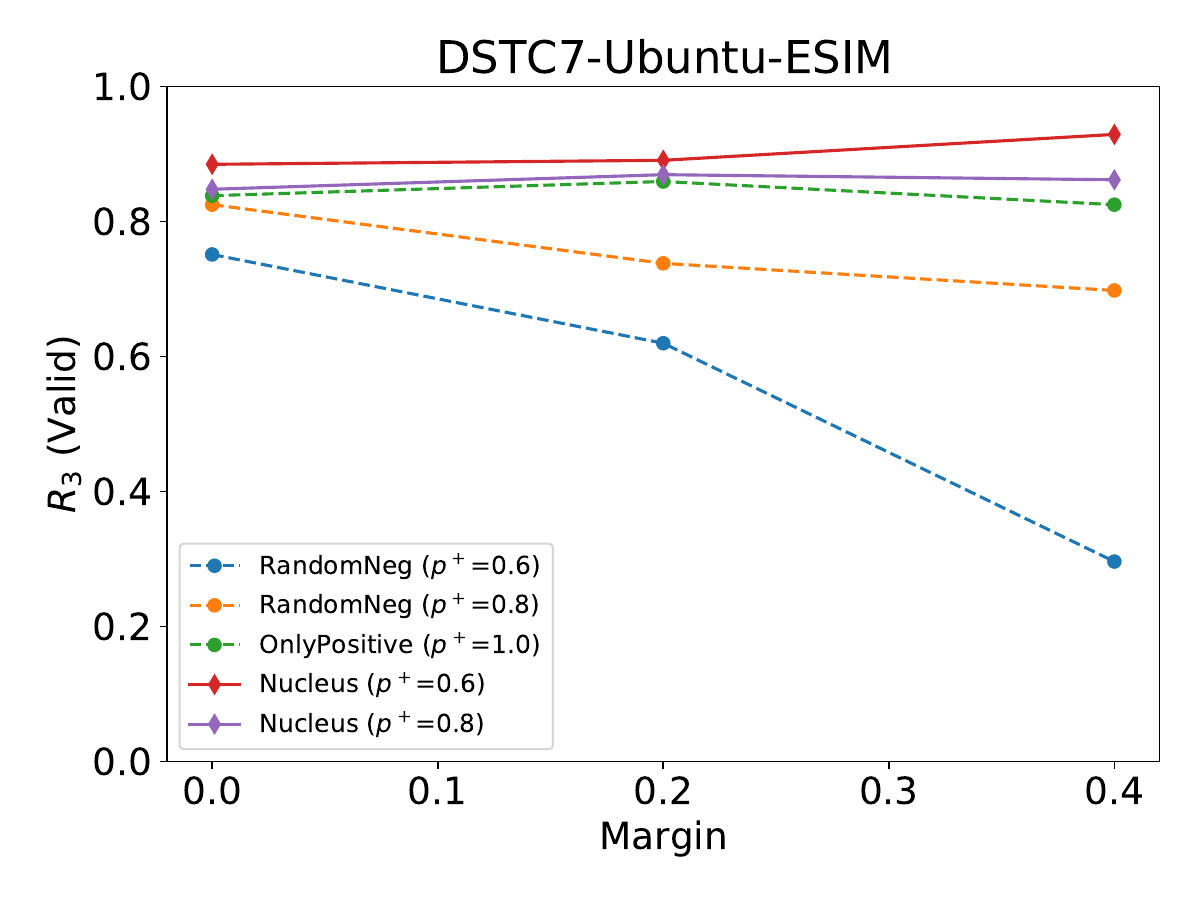}
         \caption{$\text{CORAL}_\text{ESIM}$ (DSTC7-Ubuntu)}
         \label{fig:dstc7-1}
     \end{subfigure}
     \hfill
     \begin{subfigure}[b]{0.48\textwidth}
         \centering
         \includegraphics[width=\textwidth]{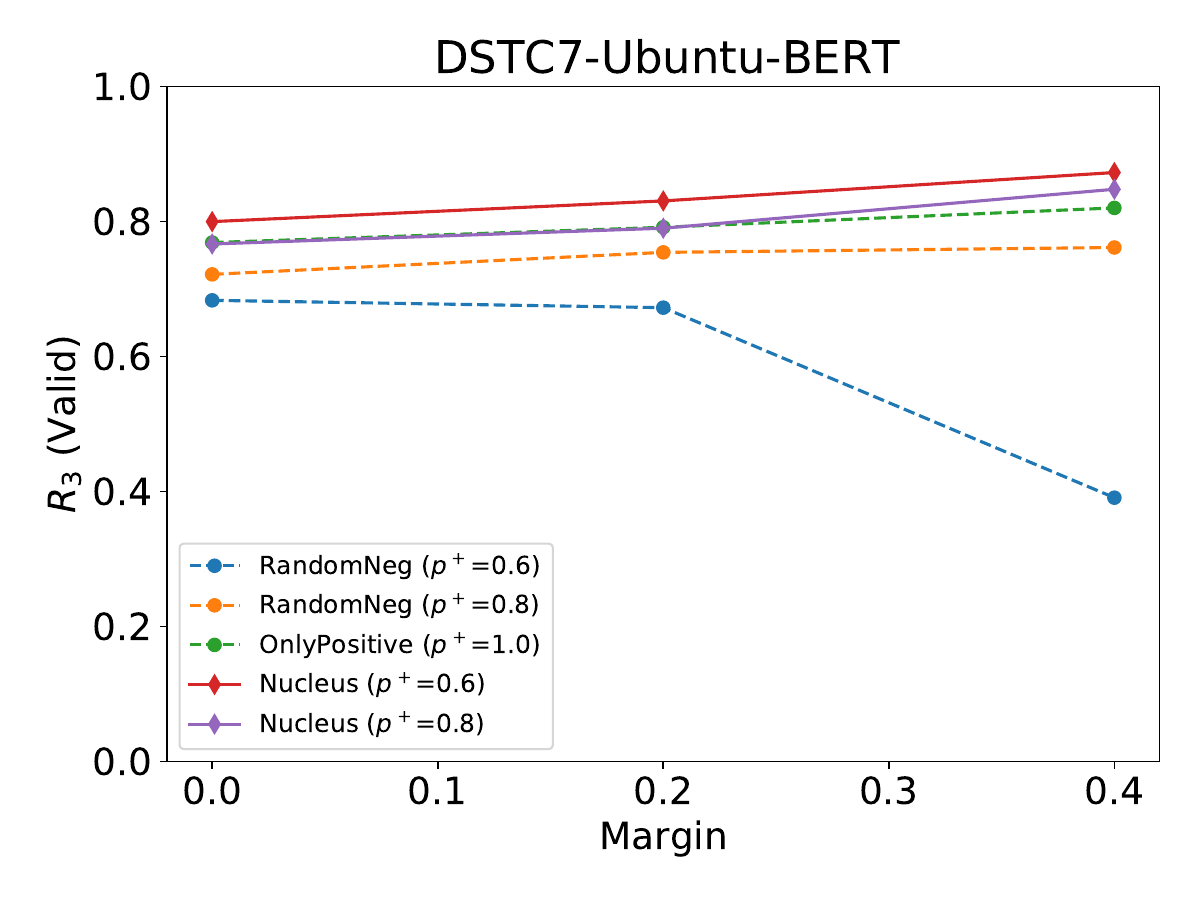}
         \caption{$\text{CORAL}_\text{BERT}$ (DSTC7-Ubuntu)}
         \label{fig:dstc7-2}
     \end{subfigure}
      \begin{subfigure}[b]{0.48\textwidth}
         \centering
         \includegraphics[width=\textwidth]{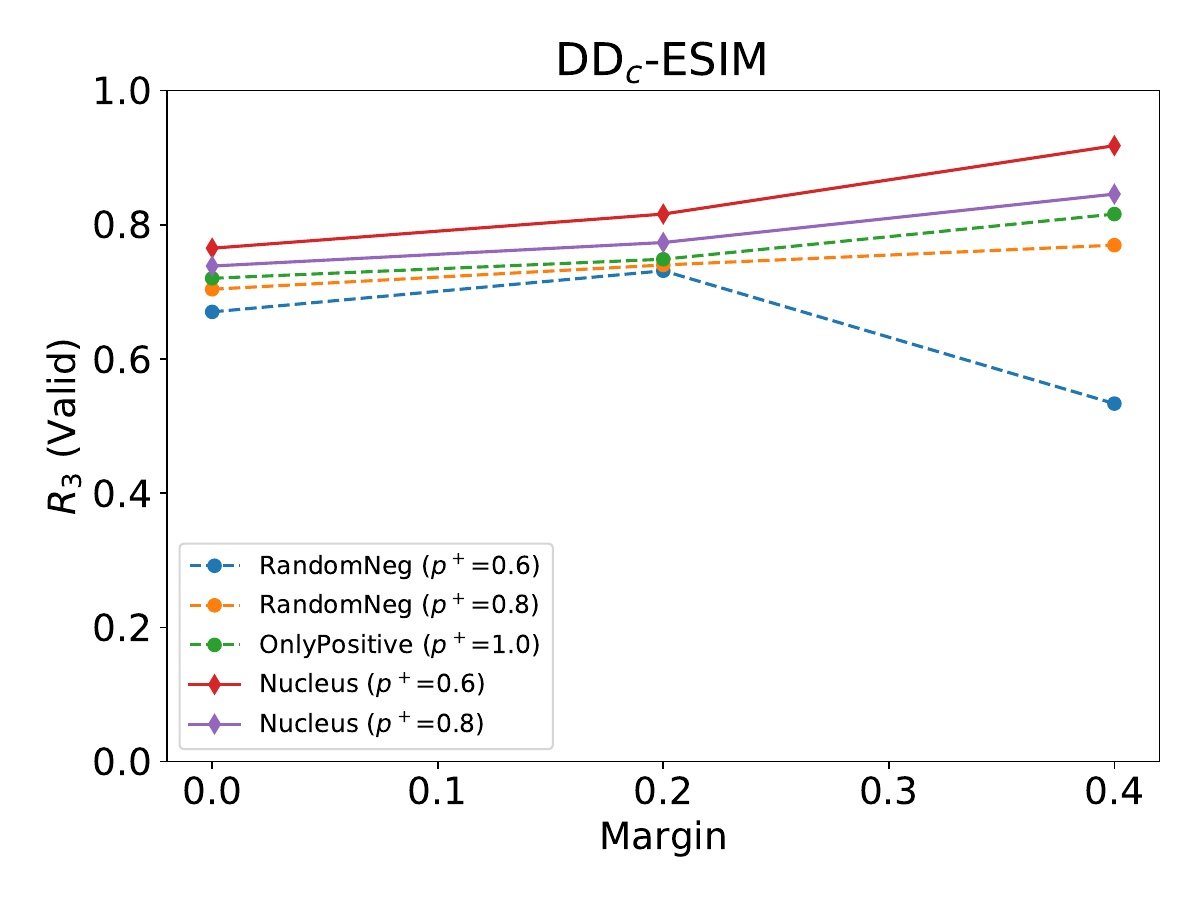}
         \caption{$\text{CORAL}_\text{ESIM}$ (DD$_c$)}
         \label{fig:ddcc1}
     \end{subfigure}
     \hfill
     \begin{subfigure}[b]{0.48\textwidth}
         \centering
         \includegraphics[width=\textwidth]{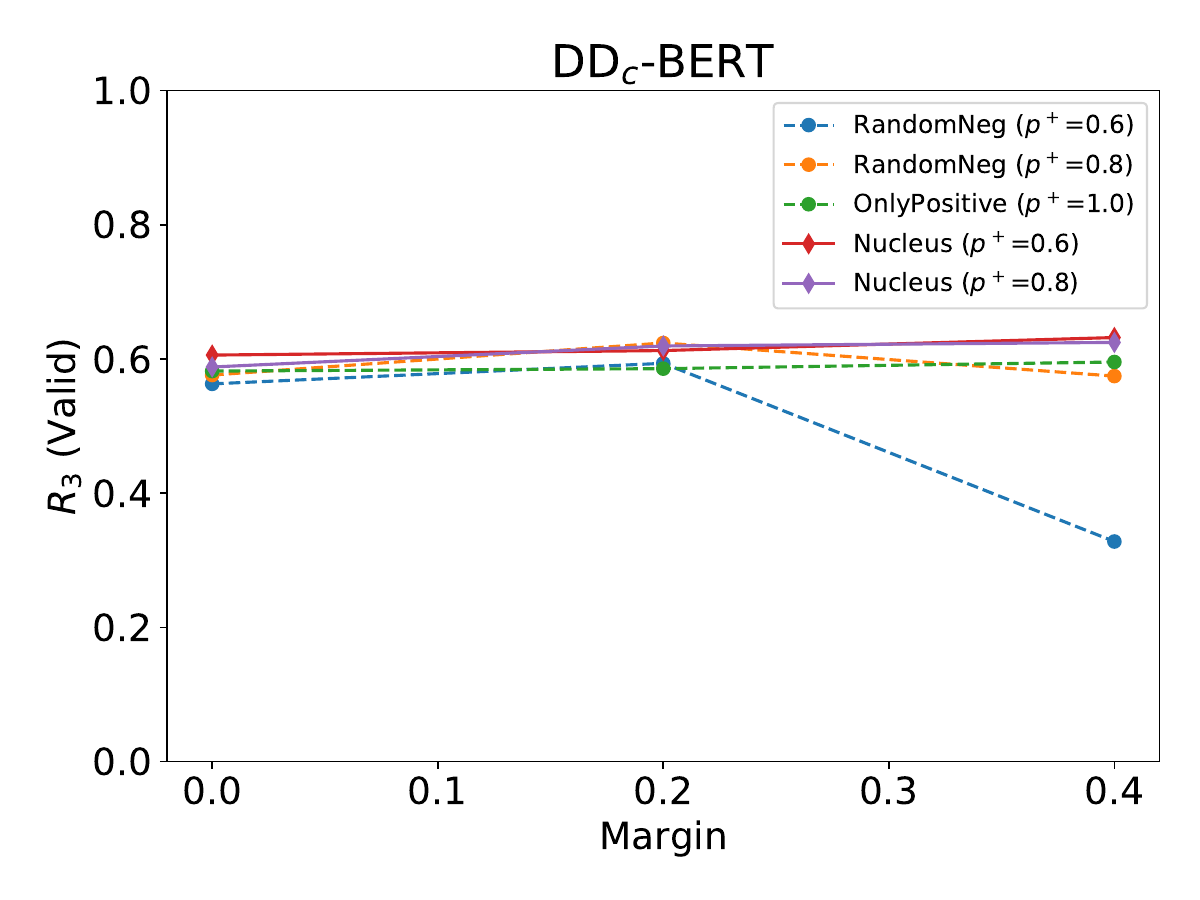}
         \caption{$\text{CORAL}_\text{BERT}$ (DD$_c$)}
         \label{fig:ddcc2}
     \end{subfigure}
     % \hfill
     % \begin{subfigure}[b]{0.48\textwidth}
     %     \centering
     %     \includegraphics[width=\textwidth]{plots/dstc7-DMI-merged-hparams.pdf}
     %     \caption{$\text{CORAL}_\text{DMI}$ (DSTC7-Ubuntu)}
     %     \label{fig:dstc7-3}
     % \end{subfigure}
    \caption{Hyperparameter Sensitivity Analysis/Ablation Studies: These plots showcase the effect of $p^+$ and margin on the final validation-$R_3$ score obtained by the corresponding CORAL model. Each lineplot corresponds to a single $p^+$ value as indicated by the legend. \textit{Note: The $R_3$ values are not comparable across any two plots.}}
    \label{fig:hparams}
\end{figure}

We conducted extensive experiments to examine the impact of different hyperparameters on the final trained model. Specifically, we varied $p^+$, $m$, and the sampling method. The comprehensive results can be found in Figure~\ref{fig:hparams}. To make comparisons, we focused on the best average reward achieved by the model on the validation set.

In general, the mix-policy setup consistently outperformed off-policy training routines. When considering the DD scenario, we noticed that lower margin values tended to yield higher $R_3$ scores. However, for DSTC7-Ubuntu and DD$_c$, the mix-policy training approach demonstrated an increase in $R_3$ as the margin value increased. Notably, we observed that mix-policy training using nucleus sampling outperformed RandomNegatives.

% \begin{figure*}
%      \centering
    
%      % \hfill
%      % \begin{subfigure}[b]{0.48\textwidth}
%      %     \centering
%      %     \includegraphics[width=\textwidth]{plots/dd_cc-DMI_BASE-merged-hparams.pdf}
%      %     \caption{$\text{CORAL}_\text{DMI}$ (DailyDial\_cc)}
%      %     \label{fig:ddcc3}
%      % \end{subfigure}
%      % \hfill     
%     \caption{Hyperparameter Sensitivity Analysis/Ablation Studies: Results for cleaned DailyDialog (DD$_c$) dataset.}
%     \label{fig:hparams-set2}
% \end{figure*}

% \section{Reward Model Evaluation}

% \enter{$R_3$ Models trained on DSTC7-Ubuntu transfers decently to DD\_c, but not the other way. This is expected given the domain-knoweldge required to solve the DSTC7-Ubuntu task.}

% \enter{See Table \ref{tab:reward-model-test}.}

% \begin{table}[!htbp]
% \centering
% \scriptsize
% \begin{tabular}{llllllll}
% \hline
%  &  & \multicolumn{3}{l}{In-Domain} & \multicolumn{3}{l}{Cross-Domain} \\ \hline
% Dataset & \textbf{R3 Model} & Recall@1 & Recall@5 & Recall@10 & Recall@1 & Recall@5 & Recall@10 \\ \hline
% DailyDialog\_CC & ESIM & 0.4629 & 0.7370 & 0.8894 & 0.4334 & 0.7037 & 0.8450 \\
%  & BERT & 0.6994 & 0.9556 & 0.9924 & 0.2171 & 0.5102 & 0.7058 \\
% DSTC7 & ESIM & 0.5720 & 0.8110 & 0.8770 & 0.2320 & 0.4090 & 0.5200 \\
%  & BERT & 0.2810 & 0.6010 & 0.7330 & 0.0460 & 0.1590 & 0.2480 \\ 
%  \hline \\
 
% \end{tabular}
% \caption{This table presents the results for the response retrieval task used for training the R3 reward model.}
% \label{tab:reward-model-test}
% \end{table}

\section{Error Analysis}

\begin{table*}[!htbp]
\caption{We did an error analysis of the generated response samples by our CORAL-BB-BERT (mixp) model on both datasets. This error analysis study was done on a set of 25 randomly selected context-response pairs from each dataset.}
\label{tab:error-analysis}
\centering
\scriptsize
\begin{tabular}{lrr}
\hline
Error Type & DD\_c & DSTC7 \\ \hline
No Errors & 19 & 16 \\
Type 1: Not Topical & 0 & 1 \\
Type 2: Agents mixed in response & 1 & 0 \\
Type 3: Wrong agent responds & 2 & 1 \\
Type 4: Generated 2 consecutive utterances & 1 & 1 \\
Type 5: Relevant but incorrect followup & 2 & 5 \\
Type 6: Relevant but incoherent to the context & 0 & 1 \\ \hline
\end{tabular}
\end{table*}

% \enter{
To gain insights into the limitations of the CORAL model and identify potential areas for improvement, we conducted an error analysis study on the responses generated by our best models (config: CORAL-BB$_\text{BERT}$-mixp). We found that, while the use of the R3 reward ensured consistent generation of contextually relevant responses, there are still a few subtle errors that our models make. These errors included difficulties in correctly attributing the next responses to user or the bot (4\% DD$_c$, 0\% DSTC7), the tendency to generate relevant but divergent follow-ups (8\% DD$_c$, 20\% DSTC7), and occasional generation of consecutive utterances as a single response (4\%). Quite notably, we observe significant variations in error distributions between the datasets (DD$_c$ and DSTC7), likely stemming from the technical nature of the DSTC7-Ubuntu dataset. We believe that these limitations appear from the current design of the $R_3$ function which focuses on topical relevance and coherence to the context. These limitations sometimes percolate down to the CORAL model trained using the $R_3$ reward function. To address these errors, we propose potential solutions such as incorporating specific self-supervision signals into the loss function during training of the reward model and utilizing better pretrained base models. We believe that implementing these measures can greatly contribute to resolving the identified limitations and improving the overall performance of the model.
% }

\section{Human Annotation Guidelines}
\begin{figure}
    \centering
    \includegraphics[width=0.7\textwidth]{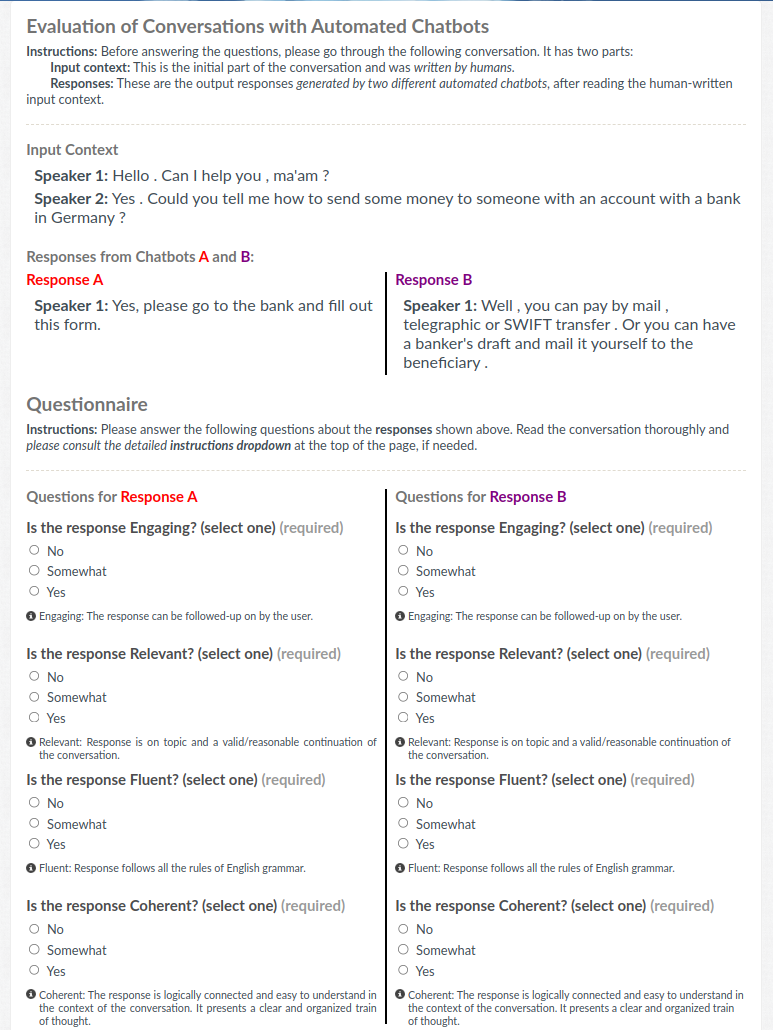}
    \caption{Web-UI used for Human Evaluation Survey. We used Appen.com for running the survey. More detailed instructions about the aspects were shown to the annotators at the start of the annotation task. The current image only shows what the annotators were able to see during task (conversation, questions and single-line tip about each of the aspects). }
    \label{fig:survey-ui}
\end{figure}

For each of the three dialog qualities, the detailed instructions and examples are shown below. These instructions were available to the workers as a dropdown item in the UI, for each question. A screenshot of the actual UI is shown in Fig. \ref{fig:survey-ui}. Following instructions were given to the annotators to help them understand the individual aspects (Engaging, Relevant, Fluent and Coherent). 

% \subsection{Task Instructions}

\subsection{Engaging}

A response is considered engaging if it can engage the user. This might be an inquisitive question or an interesting response that can be followed-up on.
\begin{itemize}
    \item No: the response is boring and does little to engage the user.
    \begin{itemize}
        \item Hi there.
\item  Oh wow! That's cool!
    \end{itemize}
    \item Somewhat: the response is not particularly engaging but still leaves room for follow-up.
    \begin{itemize}
    \item My favourite colour is blue.
\item Nope. I'm not very good with cooking.
    \end{itemize}
    \item Yes: the response is actively engaging the user and trying to move forward the conversation.
        \begin{itemize}
        \item I have a feeling that if you can dodge a wrench you can dodge a ball.
\item What kind of shows do you like?
    \end{itemize}
    
\end{itemize}

\subsection{Relevant}

A response is considered relevant if it is on topic and a valid/reasonable continuation of the conversation. An irrelevant response is one that is completely off topic, or abruptly tries to change the topic of conversation.

\begin{itemize}
\item No: the response is irrelevant. It is completely off topic or abruptly tries to change the topic of conversation.
\begin{itemize}
\item User: I think I'm starting to feel sick actually System: My favorite movie is Parasite
\item User: Can you play dodge ball? System: What kind of shows do you like?
\end{itemize}
\item Somewhat: the response is a valid continuation of the conversation, but is somewhat generic or tangential to the main topic. For example, it begins with a generic acknowledgement of the topic then changes topics.
\begin{itemize}
\item User: Have you tried seafood pancake? System: It's okay. I like beef bulgogi, but I'm not a huge fan of kimchi.
\item User: It's going great, I just finished reading '1001 inventions that changed the world' System: That sounds interesting. My favorite movie is Parasite.
\end{itemize}
\item Yes: the response is relevant, on topic and a valid continuation of the conversation.
\begin{itemize}
\item User: lol... wanna join us for dodge ball? System: Sounds like fun! I'm in!
\item User: I think I like most foods. I really like steak System: Steak is good. I especially love it when it's well done.
\end{itemize}
\end{itemize}

\subsection{Fluent}

A response is considered fluently written if it follows all the rules of English grammar and is coherent.

\begin{itemize}
    \item No: the response is not fluently written.
    \begin{itemize}
        \item System: Therefore you are. Still how long? I don't know if I can.
    \item System: OK let's do it. Pardon me? Why it shouldn't be?
    \end{itemize}
\item Somewhat: the response is somewhat fluent, but makes a minor mistake or is confusing.
        \begin{itemize}
        \item System: I am 18 of your Earth years.
\item System: Glad to hear it. What's new? Oh. Maybe you should get some rest?
    \end{itemize}
        \item Yes: the response is written fluently.
        \begin{itemize}
        \item System: I have a feeling that if you can dodge a wrench you can dodge a ball.
\item System: The Last Airbender, and Arrested Development are my favorites.
    \end{itemize}
\end{itemize}
%https://docs.google.com/document/d/1k8YLkNu65oxVpLPq3mdUZNiNGQD2uwH95m1er2m2-20/edit#

\subsection{Coherent}

A response is considered coherent if it is logically connected and easy to understand in the context of the conversation. It presents a clear and organized train of thought.

\begin{itemize}
    \item No: the response is not logical or does not make sense in the context of the conversation.
    \begin{itemize}
        \item I like pizza. The sky is green.
    \end{itemize}
    \item Somewhat: the response is somewhat logical and somewhat easy to understand, but it could be improved.
    \begin{itemize}
        \item I'm not sure what you mean. Can you explain? I like pizza.
    \end{itemize}

    \item Yes: the response is logical and easy to understand, and it presents a clear and organized train of thought.
    \begin{itemize}
        \item I think that what you're saying makes a lot of sense, and I agree with your point of view. Can you tell me more about your experience? 
    \end{itemize}
\end{itemize}

\section{Results on original DD dataset}
In Table~\ref{tab:results-dd-appendix}, we present the automatic evaluation results for response generation using non-pretrained, zero-shot and finetuned baselines and our proposed models separately for the original DD  dataset. For our proposed models, we present variants based on (1) reward function (BERT/ESIM), (2) sample generation method (off-policy and mix-policy), (3) Random or Blenderbot initialization. 

For our proposed CORAL models, %we first trained the model on the DD and DSTC7 datasets and saved the best checkpoint based on the average $R_3$ reward on the validation set. Further, 
we found $m$=0.2 and $p^+$=0.6 for CORAL$_\text{ESIM}$ and $m$=0 and $p^+$=0.6 for CORAL$_\text{BERT}$ as the best hyperparameters. Fig.~\ref{fig:hparams} shows sensitivity analysis for $p_+$ and $m$. Similar to DSTC7 and DD$_c$ datasets, CORAL-based models outperform baselines by significant margins. 

\begin{table*}[!thb]
\caption{Results for DailyDialog (DD) dataset: From the results, we can see that by optimizing the contextual $R_3$ score directly, using REINFORCE, the CORAL model is able to produce contextually relevant responses (as evident from MaUde and DEB). The average length is reported to make sure that the model is not resorting to short utterances, such as ``I don't think I know about \textit{[topic\_word]}'', just to be coherent. %Among the baselines, Blenderbot-3B, which is a large-scale pretrained model, produces the most human-like responses as per MaUde but not as diverse as the CORAL-based models. 
$\text{CORAL}_x$ denotes a Seq2Seq model trained with CORAL loss. `x' identifies the retrieval model used for the $R_3$ reward. %Note that we used the DialoGPT-medium, DialogRPT-medium, Blenderbot checkpoints. \textit{We have kept these large-scale baselines as a separate group, in the table.} Each value reported for CORAL models is computed as average of 5 runs.
} 
\label{tab:results-dd-appendix}
\centering
\scriptsize
% \resizebox{\textwidth}{!}{%
\begin{tabular}{@{}p{0.3in}lclrrrrrrrrr@{}}
    \hline
            &Model & PT?&Size& Len & BLEU & METEOR & Dist-1 & Dist-2 & MaUde$_\text{ESIM}$ & MaUde$_\text{BERT}$ & DEB(r) & DEB(a) \\ \hline
    &Ground Truth && &12.0&0.996&0.986&0.068&0.406&0.818&0.860&0.902&0.938 \\ \hline
        \multirow{3}{0.3in}{(A) No pre-\\ training}&Mirror &\xmark& 240M&7.9&0.062&0.054&0.037&0.149&0.622&0.500&0.655&0.812 \\
    &Adalabel&\xmark &90M&11.6&0.121&0.088&0.038&0.229&0.581&0.489&0.673&0.741 \\
    &ALDgen&\xmark & 68M&12.5&0.084&0.068&0.025&0.201&0.452&0.294&0.539&0.614 \\ \hline
    \multirow{3}{0.3in}{(B) Zero-\\ shot}&DialoGPT (ZS)&\cmark& 345M &8.9&0.067&0.045&0.042&0.168&0.633&0.602&0.848&0.894\\
    &DialogRPT (ZS)&\cmark & 345M& 15.9&0.081&0.053&0.027&0.119&0.652&0.606&0.854&0.871 \\
    &Blenderbot (ZS)&\cmark& 365M & 17.3&0.105&0.077&0.022&0.098&0.636&0.632&0.964&0.965 \\ \hline
    \multirow{3}{0.3in}{(C) Fine-\\ tuned}&DialoGPT (FT)&\cmark& 345M &6.1&0.075&0.068&\textbf{0.067}&\textbf{0.296}&0.788&0.783&0.858&0.915 \\
    &DialogRPT (FT)&\cmark& 345M &16.5&0.107&0.098&0.035&0.179&0.809&0.832&0.934&0.948 \\
    &Blenderbot (FT)&\cmark& 365M & 30.7&0.114&0.108&0.036&0.198&0.822&0.825&0.968&0.950 \\ \hline
   \multirow{8}{0.3in}{(D) Our \\Models} &CORAL$_\text{ESIM}$ (offp)&\xmark& 93M &10.0&0.191&0.173&0.046&0.284&0.740&0.647&0.825&0.871\\
&CORAL$_\text{ESIM}$ (mixp)&\xmark& 93M &11.0&0.212&0.197&0.043&0.258&0.758&0.659&0.831&0.881\\
&CORAL$_\text{BERT}$ (offp)&\xmark& 93M & 9.7&0.184&0.166&0.046&0.290&0.728&0.653&0.813&0.865 \\
    &CORAL$_\text{BERT}$ (mixp)&\xmark& 93M & 10.5&\textbf{0.224}&\textbf{0.208}&0.045&0.276&0.742&0.669&0.816&0.870 \\ 
 &CORAL-BB$_\text{ESIM}$ (offp)&\cmark& 365M &21.2&0.128&0.112&0.037&0.197&0.904&0.871&0.988&\textbf{0.985}\\
&CORAL-BB$_\text{ESIM}$ (mixp)&\cmark& 365M&20.5&0.129&0.111&0.037&0.194&\textbf{0.925}&0.895&\textbf{0.989}&0.984\\
    &CORAL-BB$_\text{BERT}$ (offp)&\cmark& 365M & 21.2&0.129&0.115&0.037&0.195&0.865&0.875&0.982&0.982 \\
    &CORAL-BB$_\text{BERT}$ (mixp)&\cmark& 365M & 21.1&0.131&0.114&0.033&0.179&0.889&\textbf{0.896}&\textbf{0.989}&0.983 \\
 \hline
\end{tabular}%
\end{table*}

\section{Ethical Considerations}

Like many other pretrained language representation models, the proposed model may also have learned patterns associated with exposure bias. Interpretability associated with the output is rather limited, hence users should use the outputs carefully. The proposed model generates possible response candidates, and does not filter out any ``problematic'' candidates. Thus, for applications, where candidate responses could be problematic, (e.g., offensive, hateful, abusive, etc.), users should carefully filter them out before using the output from our model.

All the datasets used in this work are publicly available. We did not collect any new dataset as part of this work.

DailyDialog: The dataset was downloaded from \url{http://yanran.li/dailydialog}. DailyDialog dataset is licensed under CC BY-NC-SA 4.0.

DSTC7-Ubuntu: The dataset was downloaded from \url{https://ibm.github.io/dstc-noesis/public/data_description.html#ubuntu}. The dataset is available under MIT license.

\section{Limitations}
% We have trained a small version of the proposed CORAL model. It will be great to see if the gains due to CORAL loss lead to similar improvements for large scale models as well.
We experimented with English datasets only. While we hope that these results will generalize to models trained on multi-lingual datasets; empirical validation needs to be done.

\end{document}